\documentclass[10pt,journal,compsoc]{IEEEtran}
\ifCLASSOPTIONcompsoc
  \usepackage[nocompress]{cite}
\else
  \usepackage{cite}
\fi

\usepackage{amsmath,amsfonts,bm}

\def\eqref#1{equation~\ref{#1}}

\def\1{\bm{1}}

\DeclareMathAlphabet{\mathsfit}{\encodingdefault}{\sfdefault}{m}{sl}
\SetMathAlphabet{\mathsfit}{bold}{\encodingdefault}{\sfdefault}{bx}{n}

\newcommand{\R}{\mathbb{R}}

\usepackage[T1]{fontenc}

\usepackage{xcolor,soul,framed} 

\colorlet{shadecolor}{yellow}
\usepackage[pdftex]{graphicx}
\graphicspath{{../pdf/}{../jpeg/}}
\DeclareGraphicsExtensions{.pdf,.jpeg,.png}

\usepackage{array}
\usepackage{mdwmath}
\usepackage{url}
\usepackage{chngcntr}
\usepackage{graphicx}
\usepackage{wasysym}
\usepackage{multirow}
\usepackage{cite}
\usepackage[normalem]{ulem}
\usepackage{amsmath}
\usepackage{amssymb}
\usepackage{booktabs}
\usepackage{setspace}
\usepackage{enumerate}
\usepackage{pifont}
\usepackage{ragged2e}
\usepackage{hhline}
\usepackage{multirow}
\usepackage{colortbl}
\usepackage{mdframed}
\usepackage{subcaption}

\definecolor{green}{RGB}{0,150,10}
\definecolor{blue}{RGB}{0,0,0}
\definecolor{red}{RGB}{0,0,0}
\definecolor{orange}{RGB}{194,153,107}

\makeatletter  
\newif\if@restonecol  
\makeatother

\usepackage[linesnumbered,ruled, vlined]{algorithm2e}
\usepackage{algpseudocode}  
\usepackage[pagebackref,breaklinks,colorlinks]{hyperref}

\begin{document}
%
\title{SPOT: Scalable 3D Pre-training via Occupancy Prediction for Learning Transferable 3D Representations}
%
%
%

\author{Xiangchao~Yan$^{\ast}$, Runjian~Chen$^{\ast}$, Bo~Zhang, Hancheng~Ye, Renqiu~Xia, Jiakang~Yuan, Hongbin Zhou, Xinyu Cai, Botian Shi, Wenqi Shao, Ping Luo, Yu Qiao, Tao Chen, and Junchi Yan
\IEEEcompsocitemizethanks{\IEEEcompsocthanksitem
Xiangchao Yan, Bo Zhang, Hancheng Ye, Hongbin Zhou, Xinyu Cai, Botian Shi, Wenqi Shao, and Yu Qiao are with Shanghai Artificial Intelligence Laboratory, Shanghai 200232, China. (Corresponding author: Bo Zhang, E-mail: zhangbo@pjlab.org.cn). \IEEEcompsocthanksitem Runjian Chen and Ping Luo are with The University of Hong Kong. \IEEEcompsocthanksitem Jiakang Yuan and Tao Chen are with School of Information Science and Technology, Fudan University. \IEEEcompsocthanksitem Renqiu Xia and Junchi Yan are with School of Artificial Intelligence, Shanghai Jiao Tong University.\IEEEcompsocthanksitem * denotes Equal Contribution.}}

%
%

\markboth{IEEE TRANSACTIONS ON Pattern Analysis and Machine Intelligence}%
{Yan \MakeLowercase{\textit{et al.}}: SPOT: Scalable 3D Pre-training via Occupancy Prediction for Learning Transferable 3D Representations}
%



\newcommand{\method}{SPOT}

\IEEEtitleabstractindextext{
\begin{abstract}
\justifying{
Annotating 3D LiDAR point clouds for perception tasks is fundamental for many applications \textit{e.g.} autonomous driving, yet it still remains notoriously labor-intensive. Pretraining-finetuning approach can alleviate the labeling burden by fine-tuning a pre-trained backbone across various downstream datasets as well as tasks. In this paper, we propose SPOT, namely Scalable Pre-training via Occupancy prediction for learning Transferable 3D representations under such a label-efficient fine-tuning paradigm. SPOT achieves effectiveness on various public datasets with different downstream tasks, showcasing its general representation power, cross-domain robustness and data scalability which are three key factors for real-world application. Specifically, we both theoretically and empirically show, for the first time, that  general representations learning can be achieved through the task of occupancy prediction. Then, to address the domain gap caused by different LiDAR sensors and annotation methods, we develop a beam re-sampling technique for point cloud augmentation combined with class-balancing strategy. Furthermore, scalable pre-training is observed, that is, the downstream performance across all the experiments gets better with more pre-training data. Additionally, such pre-training strategy also remains compatible with unlabeled data. The hope is that our findings will facilitate the understanding of LiDAR points and pave the way for future advancements in LiDAR pre-training.
}
\end{abstract}

}
\maketitle

\IEEEdisplaynontitleabstractindextext

%
\IEEEpeerreviewmaketitle

\section{Introduction}
\label{sec:intro}

\IEEEPARstart{L}{ight} Detection And Ranging (LiDAR), which emits and receives laser beams to accurately estimate the distance between the sensor and objects, serves as one of the important sensors in outdoor scenes, especially for autonomous driving. The return of LiDAR is a set of points in the 3D space, each of which contains location (the XYZ coordinates) and other information like intensity and elongation. Taking these points as inputs, 3D perception tasks like 3D object detection and semantic segmentation aim to predict 3D bounding boxes or per-point labels for different objects including cars, pedestrians, cyclists, and so on, which are important prerequisites for downstream tasks including motion prediction~\cite{pmlr-v164-jia22a,pmlr-v205-jia23a,raljia,Jia2022HDGTHD,jia2024amp} and path planning~\cite{wu2022trajectory,jia2023think,jia2023driveadapter,yang2023survey,li2024think} to achieve safe and efficient driving.

\begin{figure*}[t]
    \centering
    \small
    \subfloat[Scalability across various datasets and tasks]{
    \begin{minipage}{0.45\linewidth}{\begin{center}
    \resizebox{\linewidth}{!}{\includegraphics{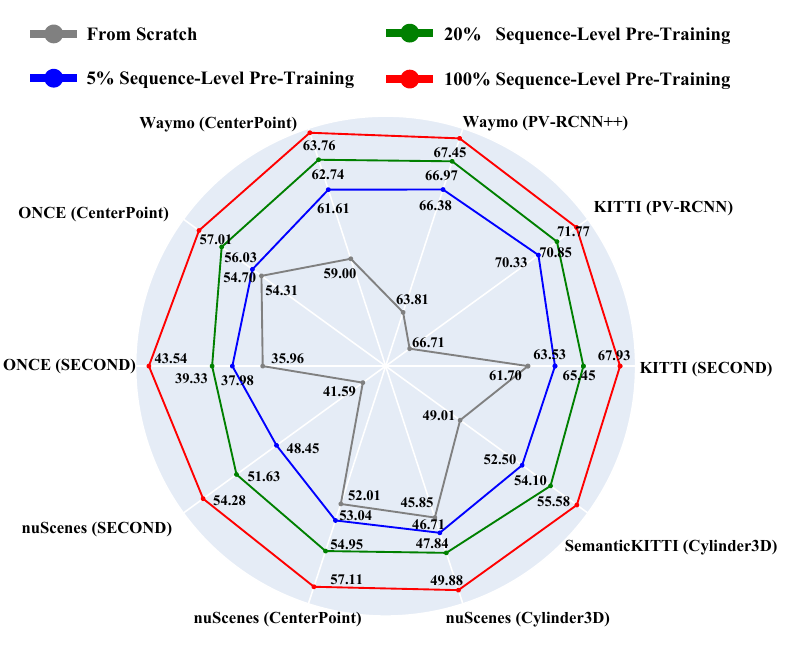}}\end{center}}\end{minipage}
    \vspace{-10pt}
    \label{fig:results_chart}}
    \subfloat[Comparison with other pre-training methods]{
    \begin{minipage}{0.48\linewidth}{\begin{center}
    \resizebox{\linewidth}{!}{\includegraphics{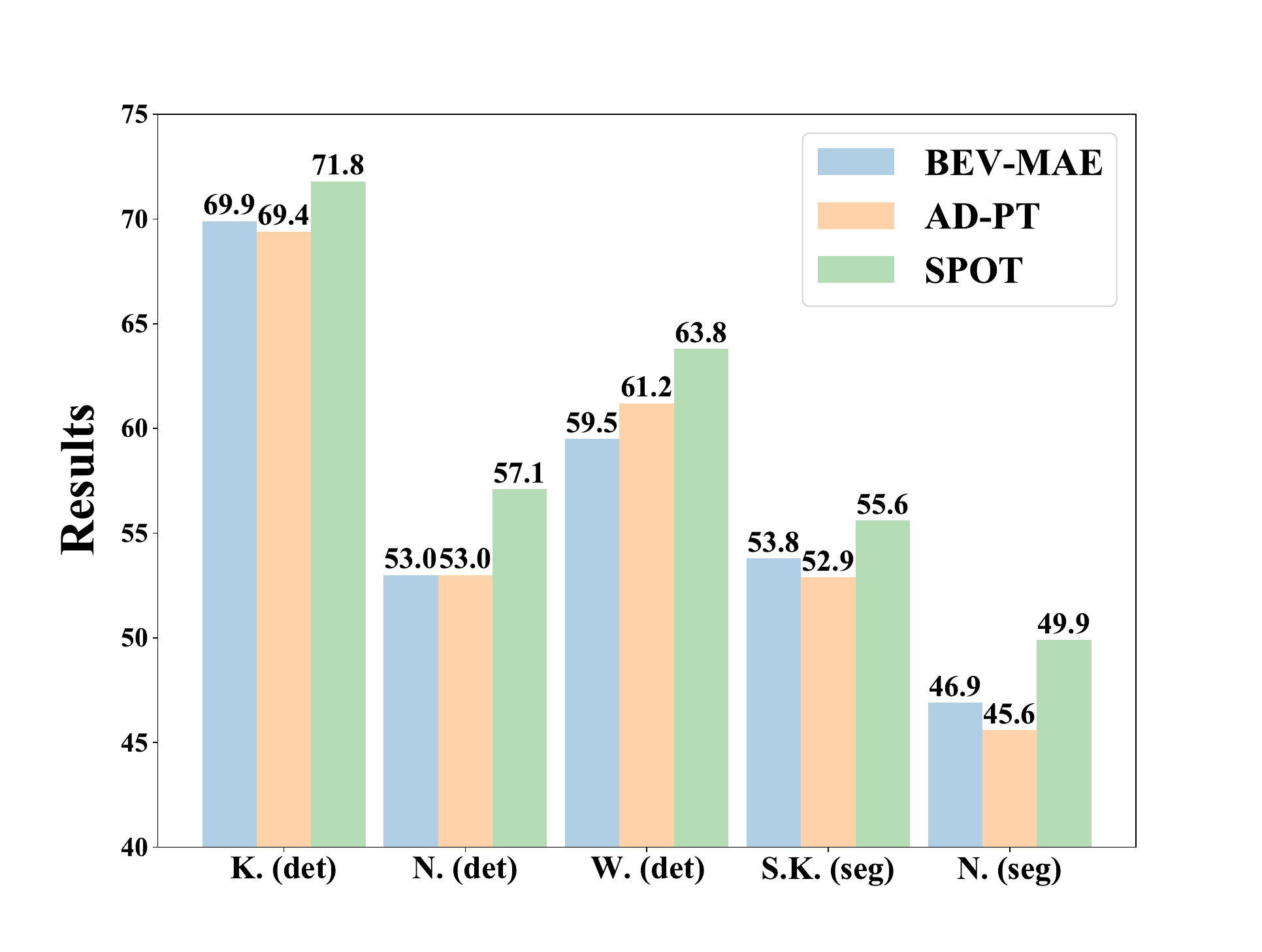}}\end{center}}\end{minipage}
    \label{fig:paradigm_comparision}}
    \caption{(a) \method\ pre-trains the 3D and 2D backbones and achieves scalable performance improvement across various datasets and tasks in label-efficient setting. Different colors indicate different amounts of pre-training data. (b) \method\ delivers the best performance on various datasets and tasks among different pre-training methods. ``\;K. (det)\;'', ``\;N. (det)\;'', ``\;W. (det)\;'' are abbreviations for KITTI, nuScenes, and Waymo detection tasks, while ``\;S.K. (seg)\;'' and ``\;N. (seg)\;'' are abbreviations for SemanticKITTI, and nuScenes segmentation tasks, respectively.}
\end{figure*}

In the past few years, research on learning-based 3D perception methods flourishes~\cite{second,centerpoint,pv-rcnn,pv-rcnn++,cylinder3d,uni3d, yuan2023bi3d}  and achieves unprecedented performance on different published datasets~\cite{kitti,semantickitti,once,nuscenes,waymo,jia2024bench2drive}. However, these learning-based methods are data-hungry and it is notoriously time-and-energy-consuming to label 3D point clouds~\cite{lu2024activead}. On the contrary, large-scale pre-training and fine-tuning with fewer labels in downstream tasks serves as a promising solution to improve the performance in label-efficiency setting. Previous methods can be divided into two streams: (1) Embraced by AD-PT~\cite{ad-pt}, semi-supervised pre-training achieves a strong performance gain when using fewer labels but limited to specific task like 3D object detection (\textbf{task-level} gap). (2) Other works including GCC-3D~\cite{gcc-3d}, STRL~\cite{strl}, BEV-MAE~\cite{bev-mae}, CO3~\cite{co3} and GD-MAE~\cite{gdmae} utilize unlabeled data for pre-training. This branch of work fails to generalize across datasets with different LiDAR sensors and annotation strategies, as shown in Fig.~\ref{fig:results_chart} (\textbf{dataset-level} gap).

To overcome both \textbf{task-level} and \textbf{dataset-level} gaps and learn general representations, we propose \method, namely \textbf{S}calable \textbf{P}re-training via \textbf{O}ccupancy prediction for learning \textbf{T}ransferable representation. {\color{red}Our key innovation lies in establishing a unified "one-for-all" pre-training paradigm that enables a single pre-training session to generalize across multiple tasks (detection, segmentation), datasets (Waymo~\cite{waymo}, KITTI~\cite{kitti}, nuScenes~\cite{nuscenes}, ONCE~\cite{once}, SemanticKITTI~\cite{semantickitti}), and sensor configurations, addressing the fundamental limitation of existing task-specific approaches.} Firstly, we argue that occupancy prediction serves as a more general pre-training task for task-level generalization, as compared to 3D object detection and LiDAR semantic segmentation. The reason lies in that occupancy prediction is based on denser voxel-level labels with abundant classes, which incorporates spatial information similar to 3D object detection as well as semantic information introduced in semantic segmentation. {\color{red}We provide the first rigorous theoretical foundation through information-theoretic analysis that justifies semantic occupancy prediction as an effective 3D pre-training task.} Besides, we consider temporally sufficient representations in the context of 3D pre-training, which contains the information shared among consecutive frames, and theoretically explain why the proposed occupancy-based pre-training outperforms self-supervised methods like MAE in downstream tasks for autonomous driving. Secondly, as the existing datasets use LiDAR sensors with various numbers of laser beams and different category annotation strategies, we propose to use beam re-sampling for point cloud augmentation and class-balancing strategies to overcome these domain gaps. Beam re-sampling augmentation simulates LiDAR sensors with different numbers of laser beams to augment point clouds from a single source pre-training dataset, alleviating the domain gap brought by LiDAR types. Class-balancing strategies apply balance sampling on the dataset and category-specific weights on the loss functions to narrow down the annotation gap. {\color{red}Beyond component integration, \method\ features several engineering insights specifically tailored for 3D pre-training scenarios, including the choice of 2D decoder over traditional 3D decoders which reduces pre-training time from 31 hours to 2.5 hours per epoch while improving parameter efficiency and downstream generalization.} Furthermore, our semi-supervised \textcolor{blue}{and weakly-supervised} pre-training experiments as described in Sec.~\ref{exp:semi_exp} provide strong evidence that \method\ consistently enhances the performance of different downstream tasks, \textcolor{blue}{
even without using any human-annotated labels}. Last but not least, we observe that using larger amounts of pre-training data leads to better performance on various  downstream tasks. This holds true even when the pre-training data are generated through pseudo-labeling. These findings indicate that \method\ is a scalable pre-training method for LiDAR point clouds, paving the way for large-scale 3D representation learning in autonomous driving.

In summary, our approach offers general representation ability, robust transferability, and  pre-training data scalabiltiy, with the specific highlights as follows:

1) We provide a theoretical analysis showing the superiority of occupancy-based pre-training task in boosting model capacity, and also empirically demonstrate the possibilities of leveraging the proposed \method\ to achieve the few-shot 3D object detection and semantic segmentation tasks. 

2) We develop a beam re-sampling augmentation combined with class-balancing strategy, which has been verified to be effective in narrowing domain gaps and boosting the model's performance across different domains.

3) Extensive experiments are conducted on few-shot 3D perception tasks and datasets including Waymo~\cite{waymo}, nuScenes~\cite{nuscenes}, ONCE~\cite{once}, KITTI~\cite{kitti}, and SemanticKITTI~\cite{semantickitti} to demonstrate the overall effectiveness of \method. As shown in Fig.~\ref{fig:paradigm_comparision}, \method\ continuously improves downstream performance as more pre-training data is used. It learns general representations and brings more consistent improvement compared to peer pre-training methods.
\section{Related Work}

\subsection{LiDAR 3D Perception} There are two main tasks on LiDAR point clouds: 3D object detection and LiDAR semantic segmentation, both of which are essential for scene understanding and control tasks. Current LiDAR 3D detectors can be divided into three main classes based on the architecture of 3D backbone in the architectures. (1) Point-based 3D detector embeds point-level features to predict 3D bounding boxes, such as PointRCNN~\cite{pointrcnn}, 3DSSD~\cite{3dssd} and PointFormer~\cite{pointformer}. (2) Voxel-based 3D detectors~\cite{voxelnet,part-a2, second, centerpoint, voxel-rcnn} divide the surrounding environment of the autonomous vehicle into 3D voxels and use sparse convolution or transformer-based encoder to generate voxel-level features for detection heads. SECOND~\cite{second} and CenterPoint~\cite{centerpoint} are popular and SOTA voxel-based 3D detectors. (3) Point-and-voxel-combined method like Fast Point R-CNN~\cite{fast_pointrcnn}, PV-RCNN~\cite{pv-rcnn}, Lidar-RCNN~\cite{lidar-rcnn} and PV-RCNN++~\cite{pv-rcnn++} utilize both voxel-level and point-level features. For the LiDAR semantic segmentation task, the goal is to predict a category label for each point in the LiDAR point clouds. Cylinder3D~\cite{cylinder3d}, the pioneering work on this task, proposes to first apply the 3D backbone to embed the voxel-level features and then a decoder for final semantic label predictions. All these methods are data-hungry and labeling for 3D point clouds is time-and-energy-consuming. To reduce the labeling burden, previous works explore semi-supervised learning~\cite{scribble,lasermix,Lim3D} and achieve excellent performance, but they are limited to specific tasks. In this work, we explore general 3D representation learning via large-scale pre-training.

\subsection{Large-scale Pre-training for Label-efficient Learning in LiDAR 3D Perception} It is promising to reduce labeling burdens by large-scale pre-training. There are two branches of methods. The first one, embraced by AD-PT~\cite{ad-pt}, is semi-supervised pre-training for 3D detection on LiDAR point cloud. AD-PT demonstrates a strong performance gain when using fewer labels. However, it suffers from limited downstream tasks (3D object detection only). The second branch of methods~\cite{xie2020pointcontrast, yin2022proposalcontrast, min2024driveworld} include GCC-3D~\cite{gcc-3d}, STRL~\cite{strl}, CO3~\cite{co3}, OCC-MAE~\cite{occ-mae}, BEV-MAE~\cite{bev-mae} and MV-JAR~\cite{mv-jar}, which utilize unlabeled data for pre-training. But these methods fail to generalize across different LiDAR sensors. In this work, we propose \method\ to pre-train the 3D backbone for LiDAR point clouds and improve performance in different downstream tasks with various sensors and architectures, as shown in Fig.~\ref{fig:paradigm_comparision}.

\begin{figure*}[t]
\centering
\includegraphics[width=0.98\linewidth]{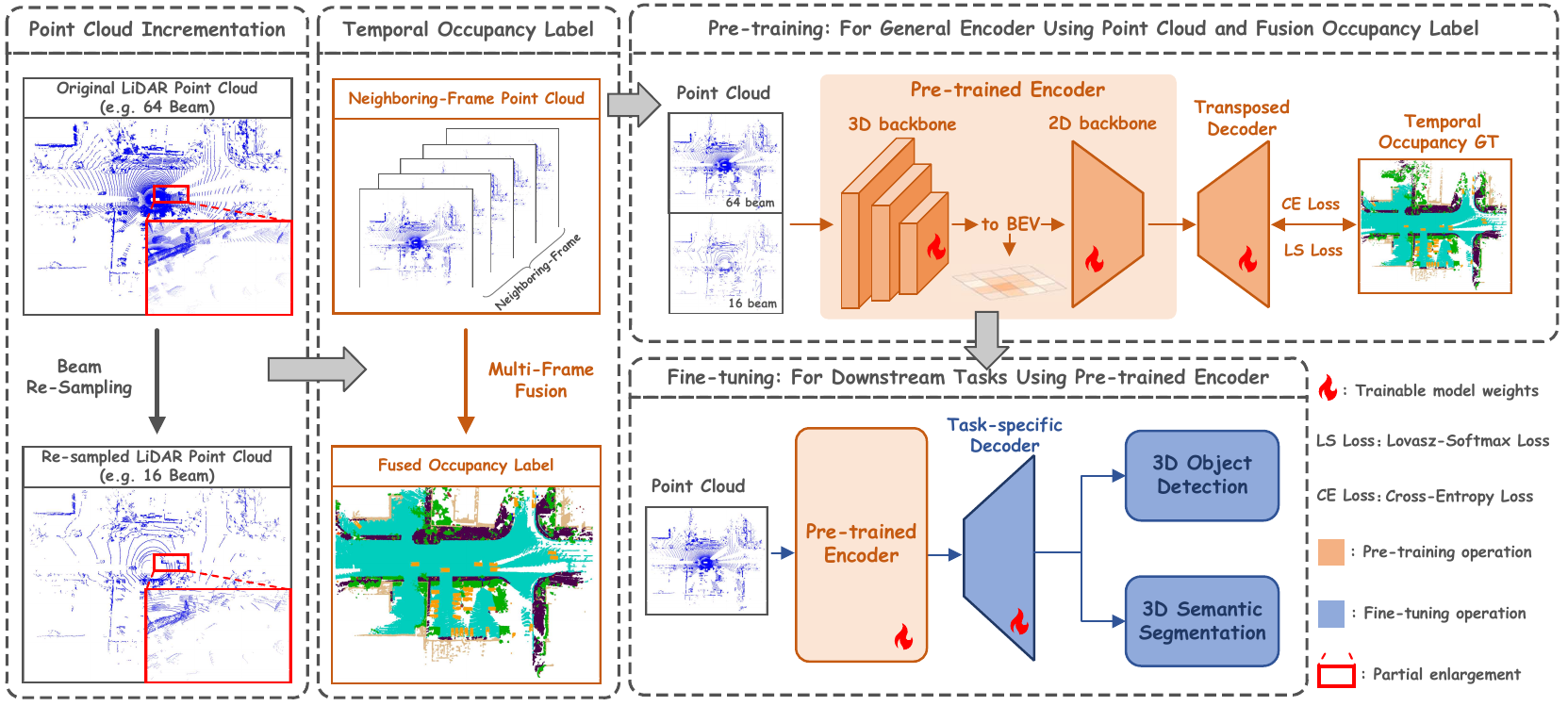}
\vspace{-4pt}
\caption{The overview of the proposed \method.  Firstly, the input LiDAR point cloud is augmented by beam re-sampling to simulate various LiDAR sensors, which helps learn general representations. Then point clouds are processed by backbone encoders consisting of 3D and 2D ones, which are utilized to initialize downstream architectures after pre-training. Next, a lightweight decoder with stacked transposed convolutions embeds the BEV features to further predict occupancy probability. Finally, we use class-balancing cross entropy loss and Lovász-Softmax loss to guide the pre-training.}
\label{fig:frame_work}
\end{figure*}

\subsection{Semantic Occupancy Prediction} The primary objective is to predict whether a voxel in 3D space is free or occupied as well as the semantic labels for the occupied ones, which enables a comprehensive and detailed understanding of the 3D environment. Represented by MonoScene~\cite{monoscene}, VoxFormer~\cite{voxformer}, TPVFormer~\cite{tpvformer}, JS3C-Net~\cite{js3c-net}, SCPNet~\cite{scpnet}, OpenOccupancy~\cite{wang2023openoccupancy}, Occformer~\cite{zhang2023occformer}, Cotr~\cite{ma2024cotr},
UniOcc~\cite{pan2023uniocc},
Pop-3D~\cite{vobecky2024pop}, SparseOcc~\cite{tang2024sparseocc}, PMAFusion~\cite{li2024pmafusion}, LowRankOcc~\cite{zhao2024lowrankocc}, and SelfOcc~\cite{huang2024selfocc}, deep learning methods achieve unprecedented performance gains on this task. For example, PMAFusion~\cite{li2024pmafusion} tries to design an effective fusion module to fuse point cloud and image features by semantic occupancy prediction. Besides, SelfOcc~\cite{huang2024selfocc} is proposed to use self-supervised 3D occupancy prediction way to learn meaningful geometric information in a 3D scene. However, these methods are specially designed for semantic occupancy prediction task and fail to learn general representations for different 3D perception tasks, such as object detection and semantic segmentation. In this paper, \method\ is proposed to use 3D semantic occupancy prediction to learn a unified 3D scene representation for various downstream tasks including 3D object detection and LiDAR semantic segmentation.
\section{The Proposed Method}

We discuss the proposed \method\ in detail. As shown in Fig.~\ref{fig:frame_work}, \method\ contains four parts: (a) Augmentations on LiDAR point clouds. (b) Encoder for LiDAR point clouds to generate BEV features, which are pre-trained and used for different downstream architectures and tasks. (c) Decoder to predict occupancy based on BEV features. (d) Loss function with class-balancing strategy. We first introduce the problem formulation as well as the overall pipeline in Sec. \ref{subsec: problem formulation and pipeline}. Then we respectively discuss beam re-sampling augmentation and class-balancing strategies in Sec. \ref{subsec: augmentations} and Sec. \ref{subsec: strategy}. In Sec.~\ref{sec:theo_anal}, \textcolor{blue}{we provide a theoretical analysis to demonstrate temporally sufficient representations for pre-training in the scenario of autonomous driving.}

\subsection{Problem Formulation and Pipeline}
\label{subsec: problem formulation and pipeline}

\subsubsection{Notation} 
To start with, we denote LiDAR point clouds $\mathbf{P}\in \R^{N\times (3+d)}$ as the concatenation of $xyz$-coordinate $\mathbf{C}\in \R^{N\times 3}$ and features for each point $\mathbf{F}\in \R^{N\times d}$, that is $\mathbf{P}=[\mathbf{C},\mathbf{F}]$. $N$ here is the number of points and $d$ represents the number of point feature channels, which is normally $d=1$ for intensity of raw input point clouds. Paired with each LiDAR point cloud, detection labels $L_{det}\in \R^{N_{det}\times 10}$ and segmentation labels for each point $L^{j}_{seg}\in \{0,1,2,...,N_{\text{cls}}\}$ ($j=1,2,...,N$) are provided. For detection labels, $N_{det}$ is the number of 3D boundary boxes in the corresponding LiDAR frame and each box is assigned $xyz$-location, sizes in $xyz$-axis (length, width and height), orientation in $xy$-plane (the yaw angle), velocity in $xy$-axis and the category label for the corresponding object. For segmentation labels, each LiDAR point is assigned a semantic label where $0$ indicates ``empty'', and $1$ to $N_{\text{cls}}$ are different categories like vehicle, pedestrian, etc.

\subsubsection{Pre-processing} 
We generate GT occupancy $\mathbf{O}\in\{0,1,2,...,N_{\text{cls}}\}^{H\times W}$ for autonomous driving pre-training following the practice in~\cite{occ3d}, where $H$ and $W$ are respectively number of voxels in $xy$-axis and Fig.~\ref{fig:frame_work} shows an example. In general, we take LiDAR point clouds in the same sequence along with their detection and segmentation labels as the inputs, and divide the labels into dynamic and static. After that, all LiDAR point clouds in that sequence can be fused to generate dense point clouds, followed by mesh reconstruction to fill up the holes. Finally, based on the meshes, we can obtain occupancy $\mathbf{O}$. For more details, please refer to~\cite{occ3d}.

\subsubsection{Encoding and Decoding} 
Given an input point cloud $\mathbf{P}\in \R^{N\times (3+d)}$, augmentations including beam re-sampling, random flip, and rotation, are first applied and result in the augmented point cloud $\mathbf{P}_{\text{aug}}\in \R^{N\times (3+d)}$. Then $\mathbf{P}_{\text{aug}}$ is embedded with sparse 3D convolution and BEV convolution backbones to obtain dense BEV features $\mathbf{F}_{\text{BEV}}\in\mathbb{R}^{\hat{H}\times\hat{W}\times\hat{d}}$ as follows:
\begin{equation}
\mathbf{F}_{\text{BEV}}=f^{\text{enc}}(\mathbf{P}_{\text{aug}}),
\end{equation}
where $\hat{H}$ and $\hat{W}$ are height and width of the BEV feature map and $\hat{d}$ is the number of feature channels after encoding. Then based on $\mathbf{F}_{\text{BEV}}$, a convolution decoder together with a Softmax operation (on the last dimension) is applied to generate dense occupancy probability prediction $\hat{\mathbf{O}}\in\R^{H\times W \times (N_{\text{cls}}+1)}$ using the following equation:
\begin{equation}
\hat{\mathbf{O}}=\text{softmax}(f^{\text{dec}}(\mathbf{F}_{\text{BEV}})),
\end{equation}
where $H$ and $W$ are the same as those of $\mathbf{O}$. For each pixel on BEV map, an $N_{\text{cls}}+1$ dimensional probability vector is predicted, each entry of which indicates the probability of the corresponding category. We observe that the decoder $f^{\text{dec}}$ should be designed to be \textbf{simple and lightweight}, allowing the encoder $f^{enc}$ to fully learn transferable representations during the pre-training process and adapt them to different downstream tasks. Therefore, it consists of only three layers of 2D transposed convolution with a kernel size of 3 and a prediction head composed of linear layers.

\subsubsection{Loss Function} To guide the encoders to learn transferable representations, a class-balancing cross-entropy loss and a Lovász-Softmax loss~\cite{lovasz} are applied on the predicted occupancy probability $\hat{\mathbf{O}}$ and the ``ground-truth'' occupancy $\mathbf{O}$. The overall loss is:
\begin{equation}
\mathcal{L} = \mathcal{L}_\text{{ce}}(\mathbf{O}, \hat{\mathbf{O}}) + \lambda \cdot \mathcal{L}_\text{{lov}}(\mathbf{O}, \hat{\mathbf{O}}),
\end{equation}
where $\lambda$ is the weighting coefficient used to balance the contributions of the two loss. For class-balancing cross-entropy loss, details are discussed in Sec. \ref{subsec: strategy}. And the Lovász-Softmax loss is a popular loss function used in semantic segmentation, whose formulation is as follows:
\begin{equation}
\begin{aligned}
\mathcal{L}_{\text{lov}}(\mathbf{O}, \hat{\mathbf{O}}) = \frac{1}{N_\text{{cls}}} \sum_{n=1}^{N_\text{{cls}}} \overline{\Delta_{J_c}}(\mathbf{M}(n)),  \\
\mathbf{M}(n)_{h,w} = \begin{cases} 1-\hat{\mathbf{O}}_{h,w,n} \ \ \ if \ n = \mathbf{O}_{h,w} \\ \hat{\mathbf{O}}_{h,w,n}  \ \ \qquad  otherwise \end{cases},
\end{aligned}
\end{equation}
where $\mathbf{M}(n)\in\R^{H\times W}$ means the errors of each pixel on BEV map of class $n$, and $h,w$ is the pixel index for the BEV map. $\overline{\Delta_{J_c}}$ denotes the Lovász extension of the Jaccard index to maximize the Intersection-over-Union (IoU) score for class $n$, which smoothly extends the Jaccard index loss based on a submodular analysis of the set function~\cite{lovasz}.

\subsection{Beam Re-sampling Augmentation}
\label{subsec: augmentations}

\begin{figure*}[t!]
    \centering
    \small
    \resizebox{0.74\linewidth}{!}{\includegraphics{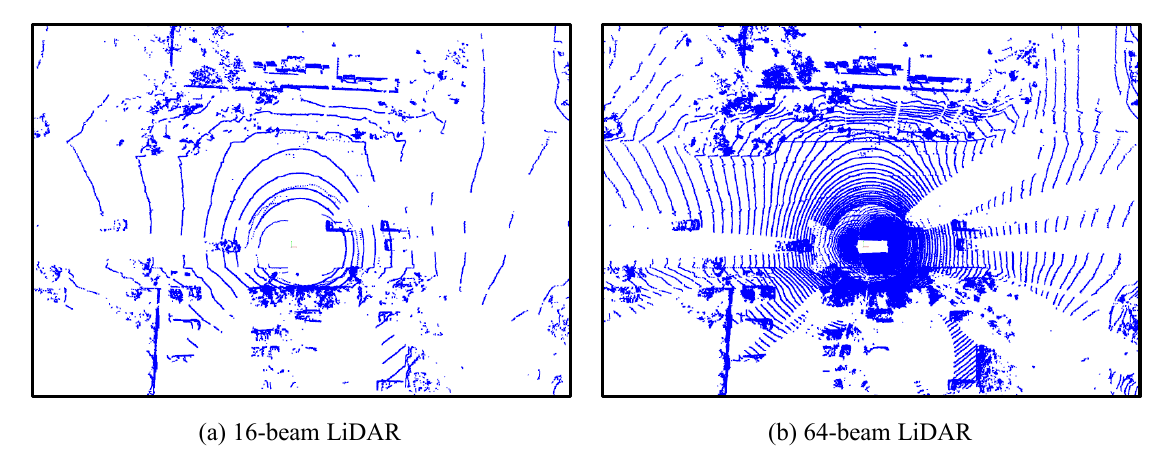}}
    \vspace{-6pt}
    \caption{Examples of different LiDAR beams}
    \label{fig: examples of LiDAR beams}
\end{figure*}

Different datasets use different LiDAR sensors to collect data. The most significant coefficient that brings domain gap is the beam numbers of LiDAR sensors, which directly determines the sparsity of the return point clouds. Fig.~\ref{fig: examples of LiDAR beams} shows an example where two LiDAR point clouds are collected by different LiDAR sensors in the same scene and it can be found that 16-beam LiDAR brings a much sparser point cloud, which results in varying distributions of the same object and degrades the performance. In order to learn general representations that benefit various datasets,  we propose equivalent LiDAR beam sampling to diversify the pre-training data.

First of all, we quantify the sparsity of point clouds collected by different LiDAR sensors. The dominant factor is beam-number and the Vertical Field Of View (VFOV) also matters. The beam density can be calculated as follows:
\begin{equation}
\begin{aligned}
\label{eq:beam_density}
    B_{\text{density}}=\frac{N_{\text{beam}}}{\alpha_{\text{up}} - \alpha_{\text{low}}},
\end{aligned}
\end{equation}
where $N_{\text{beam}}$ is the number of the LiDAR beam, and $\alpha_{\text{up}}$ and $\alpha_{\text{low}}$ respectively represent the upper and lower limits of the vertical field of view of the sensor.
 
Next, by dividing $B_\text{{density}}$ of different downstream datasets with that of the pre-training dataset, we compute re-sampling factors $R_{\text{sample}}$. Re-sampling is conducted for the pre-training data according to different $R_{\text{sample}}$. Specifically, given the original LiDAR point cloud, we transform the Cartesian coordinates $(x,y,z)$ of each point into the spherical coordinates $(r,\phi, \theta)$, where $(r,\phi, \theta)$ are the range, inclination and azimuth, respectively. Finally, uniform re-sampling is conducted on the dimension of inclination. The transformation function can be formulated as follows:
\begin{equation}
\begin{aligned}
\label{eq:LiDAR2range}
    r=\sqrt{x^2+y^2+z^2}, \\
    \phi=arctan(x/y), \\ 
    \theta=arctan(z/\sqrt{x^2+y^2}).
\end{aligned}
\end{equation}

\subsection{Class-balancing Strategies}
\label{subsec: strategy}
The contribution to downstream tasks of different categories varies. First, different datasets have various distributions over categories, which causes domain gaps and hinders learning general representations. Also, in 3D detection task, foreground classes like vehicle, pedestrian and cyclist are more important than background categories including pavement and vegetation. Thus, we propose class-balancing strategies respectively on the dataset and loss function to narrow the domain gaps.

\begin{figure}[t!]
    \centering
    \small
    \resizebox{1.0\linewidth}{!}{\includegraphics{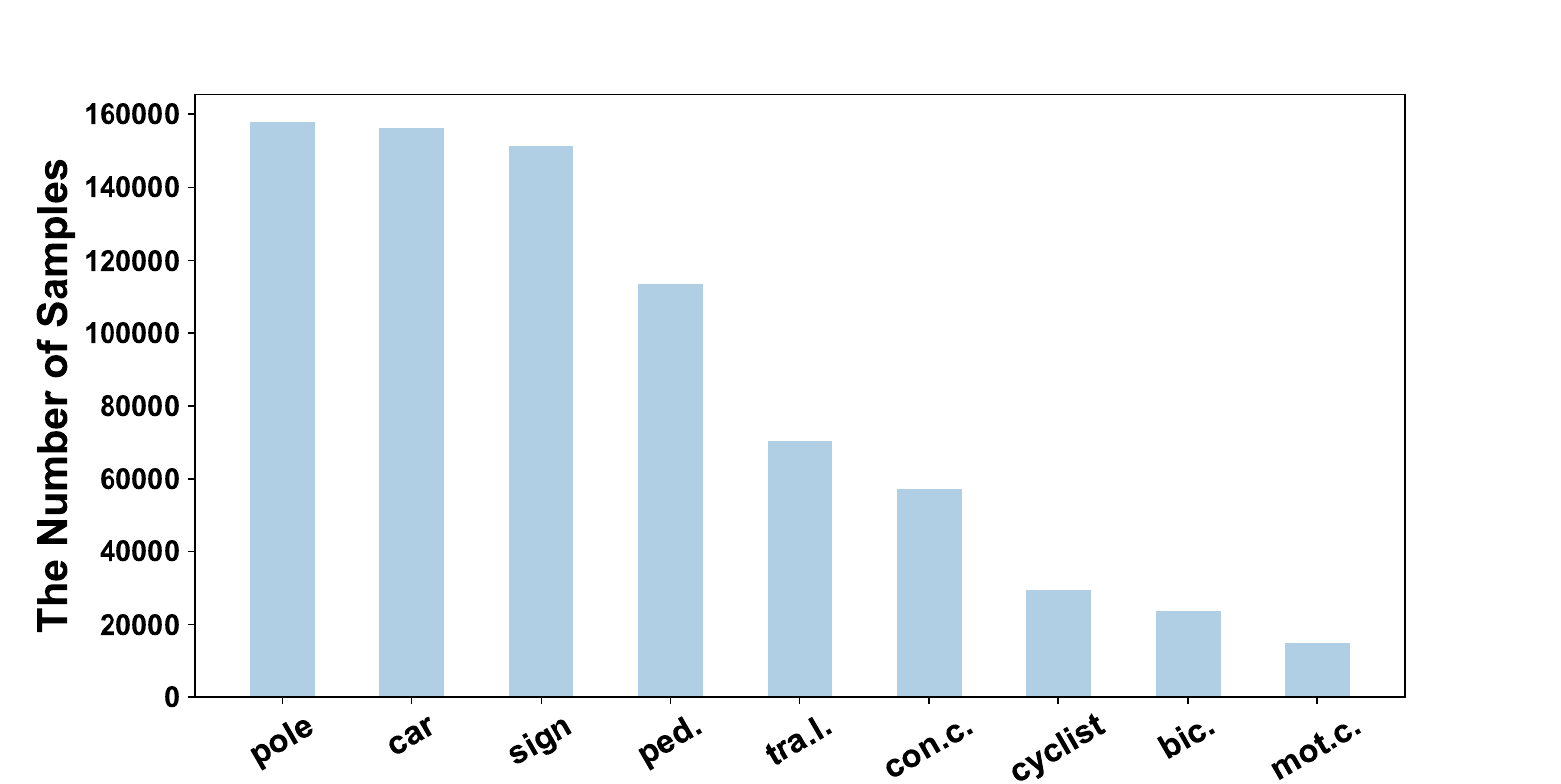}}
    \caption{Distribution of different classes.}
    \label{fig: dataset statistical analysis}
\vspace{-2pt}
\end{figure}

\subsubsection{Dataset Balancing} Considering that background classes are almost ubiquitous in every scene, we focus solely on the foreground classes in the dataset, such as cars, pedestrians, cyclists and so on. As shown in Fig.~\ref{fig: dataset statistical analysis}, we conducted a statistical analysis of the distribution of foreground semantic classes in the pre-training dataset, and it is evident that the pre-training dataset has a severe class imbalance problem. Inspired by~\cite{class-balancing-sampling}, we employ a frame-level re-sampling strategy to alleviate the severe class imbalance. Assuming that there are $N_{\text{fg}}$ foreground classes, we calculate the class sampling weights $s_{i}$ ($i=1,2,...,N_{\text{fg}}$) for each class based on the proportion of samples:
\begin{equation}
\label{eq:dataset_rebalance}
    s_{i} = \sqrt{m /{n_i}}, \ \ m = \frac{1}{N_{\text{fg}}}, \ \ n_i = \frac{N_i}{\sum_{j=1}^{N_\text{fg}}N_j},
\end{equation} 
where $N_{i}$ is the number of samples for the $i^{th}$ class. Fewer samples in a category brings higher weight $s_{i}$ for it. {\color{red}The square-root rebalancing formula is designed to provide a balanced approach that addresses class imbalance while preventing excessive duplication of rare categories.} 

{\color{red}To implement this strategy, we perform frame-level random duplication to maintain spatial-temporal consistency. Specifically, for each underrepresented category, we identify scenes containing instances of that category and apply weighted random sampling with replacement based on the computed weights $s_i$. When a frame is selected for duplication, all associated annotations are duplicated together to preserve coherence. This frame-level duplication strategy enables more effective learning of general scene representations during pre-training, thereby facilitating improved performance on downstream tasks.}


\subsubsection{Loss Function Balancing} In real-world scenarios, the surrounding 3D space of the autonomous vehicle is dominated by unoccupied states or background information. This can be harmful to the training process because the loss would be overwhelmed by a substantial amount of useless information. To overcome this challenge, we propose to assign different weights to different categories. Specifically, we assign weight $w_{\text{fg}}=2.0$ to common foreground categories including car, pedestrian, cyclist, bicycle, and motorcycle. Meanwhile, other background categories like vegetation and road are assigned $w_{\text{bg}}=1.0$ and $w_{\text{empty}}=0.01$ for unoccupied voxels.
\subsection{Theoretical Analysis}
\label{sec:theo_anal}

In this section, we borrow and extend the idea in~\cite{rethinking} to theoretically explain why the proposed occupancy-based pre-training benefits more than self-supervised pre-training methods (\textit{e.g.,} MAE) in downstream tasks of autonomous driving. First, different from the sufficient representation defined for image-level contrastive learning in~\cite{rethinking}, we consider \textbf{temporally sufficient representations} for pre-training in the scenario of autonomous driving, which contains the information shared among consecutive frames. Then, in order to present more clearer analysis, we simplify downstream tasks into classification and regression problems and present analysis on both tasks to indicate the superiority of the proposed occupancy-based pre-training. 

In the following derivation, we denote $\mathbf{O}_{\omega}^{t}$ as the occupancy map of the $t$-th frame, where $\omega$ denotes the annotation frequency of the dataset, (\textit{e.g.}, 2Hz for Waymo). The annotation of $\mathbf{O}_{\omega}^{t}$ can be formulated as $\mathbf{O}_{\omega}^{t}=\psi(\{\mathbf{P}^{t}\}_{t}, \mathbf{O}_{\omega}^{t-1})$, where $\mathbf{P}^t$ denotes the $t$-th frame points and $\psi$ represents the transformation to occupancy map from multi-frame input cloud points and the annotations of keyframes, including the multi-frame aggregation, KNN labeling and mesh reconstruction \cite{occ3d}. The representation learned from $\mathbf{O}_{\omega}^{t}$ is denoted as $z_{\text{occ}}^{t}$, while the representation learned from $\mathbf{P}^{t}$ (known as self-supervised learning, taking MAE \cite{bev-mae} as an example), is denoted as $z_{\text{mae}}^t$. The downstream task label is denoted as $T$.

\noindent\textbf{Definition 1.} \textit{(Temporally Sufficient Representation) The representation $z_{1, \text{suf}}^t$ of the $t$-th frame is temporally sufficient for another task $y_2^t$ \textbf{if and only if }$I(z_{1, \text{suf}}^t, y_2^t) = I(y_1^t, y_2^t)$, where $z_{1, \text{suf}}^t$ is learned from $y_1^t$, and $y_1^t$, $y_2^t$ are the $t$-th frame labels of two different prediction tasks that contains the shared information of consecutive frames.}

\noindent\textbf{Definition 2.} \textit{(Temporally Minimal Sufficient Representation) The representation $z_{1, \text{min}}^t$ of the $t$-th frame is temporally minimal sufficient \textbf{if and only if }$I(z_{1, \text{min}}^t, y_2^t) = \min_{z_{1, \text{suf}}^t} {I(z_{1, \text{suf}}^t, y_2^t)}$.}

\noindent\textbf{Lemma 1.} $z_{\text{occ}}^{t}$ provides more information about the downstream task $T$ than $z_{\text{mae}}^t$. That is, $I(z_{\text{occ}}^t, T) \geq I(z_{\text{mae}}^t, T)$.

\textit{Proof.} According to the paper \cite{rethinking}, single-frame MAE as a reconstruction task, can help learn an image-level sufficient representation distribution. However, since the MAE label $\mathbf{P}^t$ only contains the sufficient information of a single frame, it holds that $I(z_{\text{mae}}^t, \mathbf{O}_{\omega}^t) \leq I(z_{\text{suf}}^t, \mathbf{O}_{\omega}^t), \forall z_{\text{suf}}^t$ that is temporally sufficient. That is, $z_{\text{mae}}^t$ is a temporally minimal sufficient representation. As for $z_{\text{occ}^t}$, it learns information from multiple frames, thus is naturally one of temporally sufficient representations. Consequently, we have the relationship between $z_{\text{mae}}^t$ and $z_{\text{occ}}^t$ as follows,
\begin{equation}
\begin{aligned}
    I(z_{\text{occ}}^t, T) &= I(z_{\text{mae}}^t, T) + [I(\mathbf{O}_{\omega}^t, T|z_{\text{mae}}^t) - I(\mathbf{O}_{\omega}^t, T|z_{\text{occ}}^t)]\\
    &\geq I(z_{\text{mae}}^t, T).
\end{aligned}
\end{equation}

The first equation indicates that the mutual information $I(z_{\text{occ}}^t, T)$ can be decomposed into the minimal mutual information $I(z_{\text{mae}}^t, T)$ and the information gap between $I(\mathbf{O}_{\omega}^t, T|z_{\text{mae}}^t)$ and $I(\mathbf{O}_{\omega}^t, T|z_{\text{occ}}^t)$, where $I(\mathbf{O}_{\omega}^t, T|z_{\text{mae}}^t)$ refers to the information about $T$ that can be observed from $\mathbf{O}_{\omega}^t$ on condition of $z_{\text{mae}}^t$. Since $\mathbf{O}_{\omega}^t$ contains more information related to $T$ by multi-frame aggregation and $I(z_{\text{mae}}^t, \mathbf{O}_{\omega}^t) \leq I(z_{\text{occ}}^t, \mathbf{O}_{\omega}^t)$, we can get $I(\mathbf{O}_{\omega}^t, T|z_{\text{mae}}^t) \geq I(\mathbf{O}_{\omega}^t, T|z_{\text{occ}}^t)$. Consequently, $I(z_{\text{occ}}^t, T) \geq I(z_{\text{mae}}^t, T)$ holds.

\noindent\textbf{Theorem 1.} The upper bound of error rates in downstream tasks (including classification and regression tasks) using temporally minimal sufficient representations are higher than that of temporally sufficient representations.

\textit{Proof.} For downstream classification, we consider the Bayes error rate \cite{statistical} to estimate the lowest achievable error of the classifier. According to the paper \cite{rethinking}, for arbitrary representations $z^t$, its Bayes error rate $P_e$ satisfies that, 
\begin{equation}
    P_e \leq 1 - \exp[-H(T)+I(z^t, T)],
\end{equation}
where $H(T)$ represents the entropy of variable $T$. Since $I(z_{\text{occ}}^t, T) \geq I(z_{\text{mae}}^t, T)$, it can be concluded that the upper-bound of $P_{e, \text{occ}}$ is smaller than that of $P_{e, \text{mae}}$. This indicates that ideally $z_{\text{occ}}^t$ is expected to achieve better performance than $z_{\text{mae}}^t$ in downstream classification tasks.

For the downstream regression task, we consider the squared prediction error \cite{rethinking} to estimate the smallest achievable error of the predictor. According to the paper \cite{rethinking}, for arbitrary representations $z^t$, its minimum expected squared prediction error $R_e$ satisfies that, 
\begin{equation}
    R_e = \alpha\cdot\exp[2\cdot(H(T)-I(z^t,T))],
\end{equation}
where $\alpha$ is a constant coefficient related to the conditional distribution of squared prediction error. Similarly, since $I(z_{\text{occ}}^t, T) \geq I(z_{\text{mae}}^t, T)$, it can be concluded that the smallest achievable error of $R_{e, \text{occ}}$ is smaller than that of $R_{e, \text{mae}}$. This indicates that ideally $z_{\text{occ}}^t$ is expected to achieve better performance than $z_{\text{mae}}^t$ in downstream regression tasks.

\section{Experiments}

The goal of pre-training is to learn general representations for various downstream tasks, datasets, and architectures. We design extensive experiments to answer the question whether \method\ learns such representations in a label-efficiency way. We introduce the employed datasets and experiment setup in Sec.~\ref{dataset_desc} and~\ref{exp:setup}, respectively, followed by main results with different baselines in Sec.~\ref{exp:results}. Then in Sec.~\ref{exp:semi_exp}, we further conduct semi-supervised and weakly-supervised pre-training experiments specifically to demonstrate the applicability of SPOT in the case of utilizing very small part of annotations, and SPOT consistently demonstrates excellent performance on downstream tasks. Finally, we provide discussions about upstream pre-training and downstream fine-tuning, ablation study and visualization results in Sec.~\ref{exp:compare_occmae}, Sec.~\ref{exp:discussion} and Sec.~\ref{append:vis}.

\subsection{Dataset Description}
\label{dataset_desc}

\subsubsection{Waymo Open Dataset} Waymo Open Dataset~\cite{waymo} is a widely used outdoor self-driving dataset, which is collected in multiple cities, namely San Francisco, Phoenix, and Mountain View, using a combination of one 64-beam mid-range LiDAR and four 200-beam short-range LiDARs. This dataset contains a total of 1150 scene sequences, which are further divided into 798 training, 202 validation, and 150 testing sequences. Each sequence spans approximately 20 seconds and consists of around 200 frames of point cloud data, with each point cloud scene covering an area of approximately $150m \times 150m$.

\subsubsection{nuScenes Dataset} nuScenes Dataset~\cite{nuscenes} is a highly utilized publicly available dataset in the field of autonomous driving. It encompasses 1000 driving scenarios collected in both Boston and Singapore, with 700 for training, 150 for validation, and 150 sequences for testing. The point cloud data is collected by a 32-beam LiDAR sensor and contains diverse annotations for various tasks, (\textit{e.g.} 3D object detection and 3D semantic segmentation).

\subsubsection{KITTI Dataset} KITTI dataset~\cite{kitti}, collected in Germany, comprises data captured by a 64-beam LiDAR. It consists of 7481 training samples and 7581 test samples, with the training set further divided into 3712 and 3769 samples for training and validation, respectively. It is worth noting that unlike other datasets, KITTI dataset only provides labels within the front camera field of view.

\subsubsection{ONCE Dataset} ONCE~\cite{once} is a large-scale autonomous dataset collected in China using a 40-beam LiDAR. It encompasses a diverse range of data collected at various times, under different weather conditions, and across multiple regions. The dataset comprises over one million frames of point cloud data, with approximately 15K frames containing annotations. The remaining unlabeled point cloud data serves as resources for weakly-supervised and semi-supervised algorithms.

\begin{table*}[t]
\vspace{0.25cm}
\setlength\tabcolsep{3pt}
    \caption{Few-shot performance of SPOT on nuScenes validation set. P.D.A. denotes Pre-training Data Amount. We fine-tune on 5\% nuScenes training data.} 
    \vspace{-0.2cm}
    \label{tab:nusc_results}
    \centering
    \begin{small}
    \resizebox{1.0\linewidth}{!}
    {
        \begin{tabular}{c | c | c | c  c | c c c c c c c c c c}
            \toprule[1pt]
            Detector & Method & P.D.A. & mAP & NDS & Car & Truck & CV. & Bus & Trailer & Barrier & Motor. & Bicycle & Ped. & TC. \\
            \cmidrule{1-15}
            \multirow{6}{*}{SECOND~\cite{second}}                 & From Scratch & - & 32.16 & 41.59 & 69.13 & 33.94 & 10.12 & 46.56& 17.97 & 32.34 & 15.87 & 0.00 & 57.30 & 37.99 \\
                        & BEV-MAE~\cite{bev-mae} & 100\% & 32.09 & 42.88 & 69.84 & 34.79 & 8.19 & 48.36 & 22.46 & 32.67 & 13.01 & 0.13 & 56.10 & 35.33 \\
                        & AD-PT~\cite{ad-pt} & 100\% & 37.69 & 47.95 & 74.89 & 41.82 & 12.05 & 54.77 & 28.91 & 34.41 & 23.63 & 3.19 & 63.61 & 39.54 \\
                        & SPOT (ours) & 5\% & 37.96 & 48.45 & 74.74 & 37.94 & 12.17 & 54.94 & 27.69 & 38.03 & 22.91 & 2.55 & 64.27 & 44.31 \\
                       & SPOT (ours) & 20\% & 39.63 & 51.63 & 75.58 & 41.41 & 12.95 & 55.67 & \textbf{29.92} & 40.13 & 23.26 & 4.77 & 70.40 & 42.18 \\
                        & SPOT (ours) & 100\% & \textbf{42.57} & \textbf{54.28} & \textbf{76.98} & \textbf{42.86} & \textbf{14.54} & \textbf{59.56} & 29.30 & \textbf{44.04} & \textbf{30.91} & \textbf{7.52} & \textbf{72.70} & \textbf{47.26} \\
            \cmidrule{1-15} 
            \multirow{6}{*}{CenterPoint~\cite{centerpoint}}                 & From Scratch & - & 42.37 & 52.01 & 77.13 & 38.18 & 10.50 & 55.87 & 23.43 & 50.50 & 35.13 & 15.18 & 71.58 & 46.16 \\
                        & BEV-MAE~\cite{bev-mae} & 100\% & 42.86 & 52.95 & 77.35 & 39.95 & 10.87 & 54.43 & 25.03 & 51.20 & 34.88 & 15.15 & 72.74 & 46.96 \\
                        & AD-PT~\cite{ad-pt} & 100\% & 44.99 & 52.99 & 78.90 & \textbf{43.82} & 11.13 & 55.16 & 21.22 &  \textbf{55.10} & 39.03 & 17.76 & 72.28 & \textbf{55.43} \\
                        &  SPOT (ours) & 5\% & 43.56 & 53.04 & 77.21 & 38.13 & 10.45 & 56.41 & 24.19 & 50.33 & 37.74 & 18.55 & 73.97 & 48.59 \\
                        &  SPOT (ours) &20\% & 44.94 & 54.95 & 78.30 & 40.49 & 12.32 & 56.68 & 28.10 & 51.77 & 35.93 & 22.46 & 75.98 & 47.38 \\
                        &  SPOT (ours) & 100\% & \textbf{47.47} & \textbf{57.11} & \textbf{79.01} & 42.41 & \textbf{13.04} & \textbf{59.51} & \textbf{29.53} & 54.74 & \textbf{42.54} & \textbf{24.66} & \textbf{77.65} & 51.65 \\
            \bottomrule[1pt]
        \end{tabular}
    }
    \end{small}
\end{table*}

\begin{table*}[tb!]
    \caption{Few-shot performance ($\text{AP}_{\text{3D}}$) of SPOT on KITTI validation set. P.D.A. represents the Pre-training Data Amount, and fine-tuning is performed on 20\% KITTI training data.} 
    \vspace{-0.2cm}
    \label{tab:kitti_results}
    \centering
    \begin{small}
    \resizebox{1.0\linewidth}{!}
    {
        \begin{tabular}{c | c | c | c | c c c | c c c | c c c }
            \toprule[1pt]
             \multirow{2}{*}{Detector} & \multirow{2}{*}{Method} & \multirow{2}{*}{P.D.A.} & mAP & \multicolumn{3}{c|}{Car} & \multicolumn{3}{c|}{Pedestrian} & \multicolumn{3}{c}{Cyclist} \\
            \cmidrule{4-13}
            & & &(Mod.) & Easy & Mod. & Hard & Easy & Mod. & Hard & Easy & Mod. & Hard \\
            \cmidrule{1-13}
            \multirow{6}{*}{SECOND~\cite{second}} & From Scratch & -  & 61.70 & 89.78 & 78.83 & 76.21 & 52.08 & 47.23 & 43.37 & 76.35 & 59.06 & 55.24 \\
            & BEV-MAE~\cite{bev-mae}  & 100\% & 63.45 & 89.50 & 78.53 & 75.87 & 53.59 & 48.71 & 44.20 & 80.73 & 63.12 & 58.96 \\
            & AD-PT~\cite{ad-pt}  & 100\% & 65.95 & 90.23 & 80.70 & \textbf{78.29} & 55.63 & 49.67 & 45.12 & 83.78 & 67.50 & 63.40 \\
            & SPOT (ours)  & 5\% & 63.53 & 90.82 & 80.69 & 77.91  & 54.82 & 50.22 & 46.38 & 80.80 & 63.53 & 59.31 \\
            & SPOT (ours)  & 20\% & 65.45 & 90.55 & 80.59 & 77.56 & 56.07 & 51.68 & 47.56 & 83.52 & 65.45 & 61.11 \\
            & SPOT (ours) & 100\% & \textbf{67.36} & \textbf{90.94} & \textbf{81.12} & 78.09 & \textbf{57.75} & \textbf{53.03} & \textbf{47.86} & \textbf{87.00} & \textbf{67.93} & \textbf{63.50} \\

             
            \cmidrule{1-13}
            \multirow{6}{*}{PV-RCNN~\cite{pv-rcnn}} & From Scratch &  - & 66.71 & 91.81 & 82.52 & 80.11 & 58.78 & 53.33 & 47.61 & 86.74 & 64.28 & 59.53 \\
            & BEV-MAE~\cite{bev-mae}  & 100\% & 69.91 & 92.55 & 82.81 & 81.68 & 64.82 & 57.13 & 51.98 & 88.22 & 69.78 & 65.75 \\
            & AD-PT~\cite{ad-pt} & 100\% &  69.43 & 92.18 & 82.75 & 82.12 & 65.50 & 57.59 & 51.84 & 84.15 & 67.96 & 64.73 \\
             & SPOT (ours) & 5\%  & 70.33  & \textbf{92.68} & 83.18 & \textbf{82.26}  & 63.82 & 56.14 & 51.12 & 89.18 & 71.68 & 67.17 \\
            & SPOT (ours) & 20\%  & 70.85  & 92.61 & 83.06 & 82.03  & 65.66 & 58.02 & 52.55 & \textbf{89.77} & 71.48 & \textbf{68.01} \\
            & SPOT (ours) & 100\% & \textbf{71.77} & 92.19 & \textbf{84.47} & 82.02 & \textbf{67.31} & \textbf{59.14} & \textbf{53.41} & {89.71} & \textbf{71.69} & {67.10} \\

             
           \bottomrule[1pt]
        \end{tabular}
    }
    \end{small}
\end{table*}

\subsubsection{SemanticKITTI Dataset} SemanticKITTI dataset~\cite{semantickitti} is a large-scale dataset based on the KITTI vision, collected by a 64-beam LiDAR sensor. It has 22 sequences, of which sequences 0-7 and 9-10 are used as the training set (19K frames in total), and sequence 8 (4K frames) is used as the validation set, and the remaining 11 sequences (20K frames) as the test set.

\subsection{Experimental Setup}
\label{exp:setup}

\begin{table*}[t]
\vspace{0.25cm}
    \centering
        \caption{Few-shot performance of SPOT on SemanticKITTI validation set for \textbf{segmentation task} using 100\% pre-training data. We fine-tune on 10\% training data and show the results of some of the categories.} 
        \vspace{-0.2cm}
        \label{tab:semtickitti_results}
        \centering
        \setlength\tabcolsep{8.0pt}
        \resizebox{0.99\linewidth}{!}
        {
            \begin{tabular}{c | c | c | c| c| c| c| c| c| c| c}
                \toprule[1pt]
                Backbone & Method & mIOU & car & truck & bus & person & bicyclist  & road  &  fence &  trunk \\
                \cmidrule{1-11} 
                \multirow{4}{*}{Cylinder3D} & From Scratch & 49.01 & 93.73 & 38.03 & 25.42 & 35.52 & 0.00 & 92.55 & 46.46 & 65.22\\    
                & BEV-MAE~\cite{bev-mae} & 53.81 & 94.06 & 58.46 & 38.13 & 50.08 & 51.46 & 92.46 & 46.96 & 62.28\\ 
                & AD-PT~\cite{ad-pt} &  52.85 & 94.02 & 42.03 & 36.90 & 50.26 & 49.49 & 91.94 & 49.90 & 60.10 \\ 
                & SPOT (ours) & \textbf{55.58} & \textbf{94.34} &\textbf{61.27} &\textbf{43.01} &\textbf{55.56} &\textbf{67.61} &\textbf{92.61} &\textbf{52.81} &\textbf{67.17}\\  
    
                \bottomrule[1pt]
            \end{tabular}
        }
\end{table*}

\subsubsection{Pre-training Dataset} 
For all downstream fine-tuning experiments, we use the \textit{Waymo Open dataset}~\cite{waymo} as our pre-training dataset. We refer to such pre-training setup as the \textbf{one-for-all setting}, meaning that the encoder only needs to be pre-trained once using the proposed SPOT and can then be deployed to downstream datasets and tasks. The advantage of one-for-all setting is that different downstream applications can load the same pre-trained checkpoint to gain performance. However, such a setting also faces serious challenges, such as domain gaps~\cite{zhang2023resimad} between different downstream datasets.

Following the methodology mentioned in Sec. \ref{subsec: problem formulation and pipeline}, we generate dense occupancy labels for each sample where $N_{\text{cls}}=15$. This means 15 semantic categories including car, pedestrian and motorcycle, as well as ``empty'' are marked for each voxel. To evaluate the scalability of \method, we partition Waymo into $5\%$, $20\%$, and $100\%$ subsets at the sequence level and perform the pre-training on different subsets.


\subsubsection{Downstream Datasets and Evaluation Metrics} 

Popular LiDAR perception tasks include 3D object detection and LiDAR semantic segmentation. For detection, we cover the vast majority of currently available datasets, including \textit{KITTI}~\cite{kitti}, \textit{nuScenes}~\cite{nuscenes} and \textit{ONCE}~\cite{once} with popular 3D detectors including SECOND~\cite{second}, CenterPoint~\cite{centerpoint} and PV-RCNN~\cite{pv-rcnn} for evaluation. \textbf{\textit{nuScenes}} covers 28,130 samples used for training and 6,019 samples used for validation. We evaluate the performance using the official Mean Average Precision (mAP) and nuScenes Detection Score (NDS)~\cite{nuscenes}. For \textbf{\textit{KITTI}}, we report the results using three levels of mAP metrics: easy, moderate, and hard, following the official settings in~\cite{kitti}. \textit{\textbf{ONCE}} contains 19k labeled LiDAR point clouds, of which 5K point clouds are used for training, 3K for validation and 8K for testing. For evaluation, we follow~\cite{once} to use the mAP metrics by different ranges: 0-30m, 30-50m, and 50m-Inf. For semantic segmentation, we conduct experiments on \textit{SemanticKITTI}~\cite{semantickitti} and \textit{nuScenes}~\cite{nuscenes} with the famous LiDAR segmentor Cylinder3D~\cite{cylinder3d}. \textbf{\textit{SemanticKITTI}} is divided into a train set with 19,130 samples together with a validation set with 4,071 frames. The evaluation metric of the two datasets adopts the commonly used mIoU (mean Intersection over Union). To compute mIoU, per-category IoU is first computed as $\text{IoU}_i=\frac{\text{TP}_i}{\text{TP}_i+\text{FP}_i+\text{FN}_i}$, where $\text{TP}_i$, $\text{FP}_i$ and $\text{FN}_i$ denote true positive, false positive and false negative for class $i$, respectively. Then IoUs for different classes are averaged to get the final mIoU.


\subsubsection{Implementation Details} 

We select two representative pre-training methods for unsupervised (BEV-MAE~\cite{bev-mae}) and supervised (AD-PT~\cite{ad-pt}) branches, respectively. For pre-training phase, we adopt commonly used 3D and 2D backbones in~\cite{second,centerpoint,pv-rcnn} and $N_{\text{cls}}=15$, $\lambda=1$. We train 30 epochs with the Adam optimizer, using the one-cycle policy with a learning rate of 0.003. For the downstream detection task, we train 30 epochs for nuScenes, 80 epochs for KITTI and ONCE. For the downstream segmentation task, we train 20 and 10 epochs for SemanticKITTI and nuScenes, respectively. Our experiments are implemented based on 3DTrans~\cite{3DTrans}, using 8 NVIDIA Tesla A100 GPUs. \textbf{Note that} our experiments are under label-efficiency setting, which means that we conduct fine-tuning on a randomly selected subset of the downstream datasets (\textit{e.g.}, $5\%$ for \textit{nuScenes} detection task, $20\%$ for \textit{KITTI} and \textit{ONCE} detection tasks, and $10\%$ for \textit{SemanticKITTI} and \textit{nuScenes} segmentation tasks). {\color{blue}{All evaluation results reported in this paper are conducted on the validation split of the respective datasets.}}

\subsection{Main Results}
\label{exp:results}

\subsubsection{nuScenes Detection} 
Equipped with different types of LiDAR sensors, the domain gap between the pre-training dataset (Waymo) and the downstream dataset (nuScenes) is non-negligible. By harnessing the capabilities of \method, which learns general 3D scene representations, it can be found in Table~\ref{tab:nusc_results} that \method\ achieves considerable improvements on the SECOND~\cite{second} and CenterPoint~\cite{centerpoint} detectors compared to other pre-training strategies. Specifically, when pre-trained by 100\% Waymo sequence-level data, \method\ achieves the best overall performance (mAP and NDS) among all the pre-training methods including randomly initialization, BEV-MAE~\cite{bev-mae} and AD-PT~\cite{ad-pt}, improving training-from-scratch by up to 10.41\% mAPs and 12.69\% NDS. Scalable pre-training can also be observed when increasing the amount of pre-training data. When further looking into the detailed categories, \method\ almost achieves the best performance among all the categories for both detectors. For example, \method\ improves  SECOND on Bus, Trail, Barriers, Motorcycle and Pedestrian for more than 10\% mAP compared to training from scratch, which is essential for downstream safety control in real-world deployment.

\subsubsection{KITTI Detection} 
Despite KITTI using the same type of LiDAR sensor as that in the Waymo dataset, KITTI only employs front-view point clouds for detection, which still introduces domain gaps. In Table \ref{tab:kitti_results}, it can be found that, SECOND~\cite{second} and PV-RCNN~\cite{pv-rcnn} detectors with \method\ method are significantly and continuously improved as more pre-training data are added. For 100\% pre-training data, the improvements are respectively 5.66\% and 5.06\% mAPs at moderate level. For detailed categories, \method\ brings consistent improvement over different classes. When we focus on the moderate level, the most commonly used metrics, \method\ achieves the best among all the initialization methods for all classes, which shows great potential in real-world applications.

\vspace{-2pt}
\subsubsection{ONCE Detection} 
\begin{figure}[t]
\vspace{0.05cm}
    \centering
    \small
    \resizebox{0.97\linewidth}{!}{\includegraphics{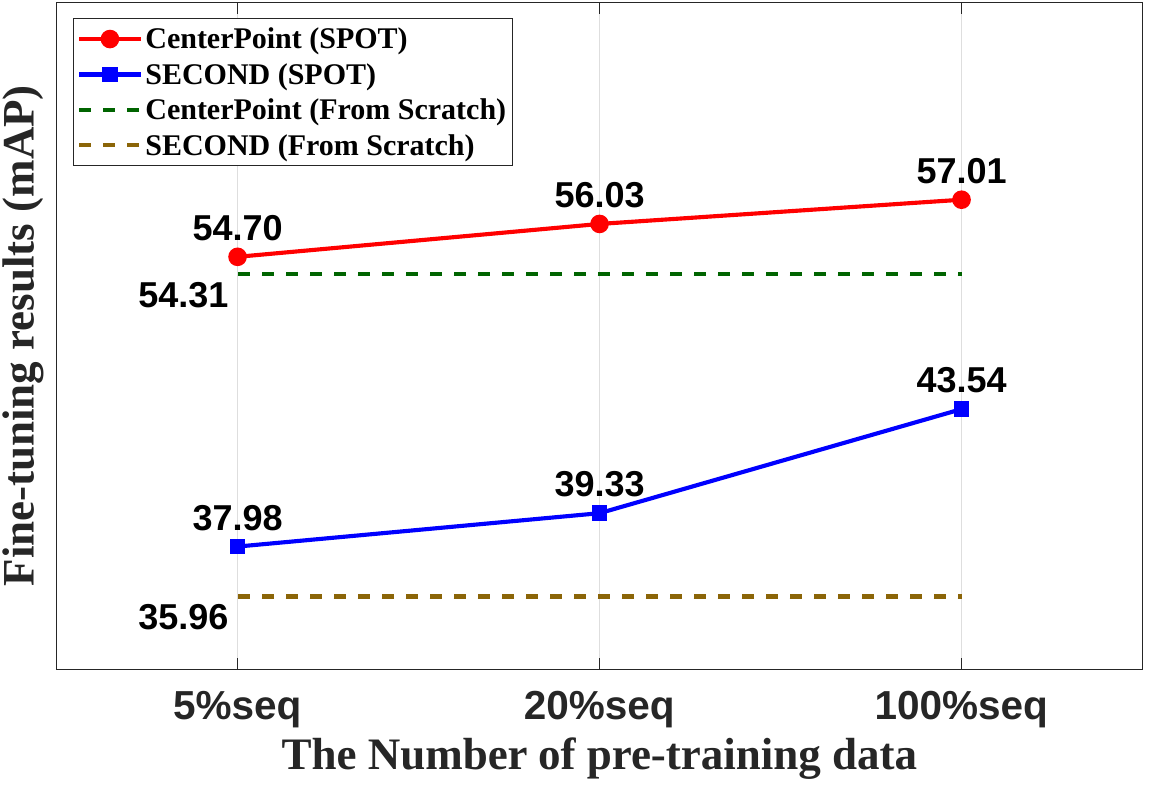}}
    \caption{Fine-tuning on ONCE validation set for detection task, where 20\% training data are used}
    \label{fig:once_results}
\end{figure}

\vspace{-1pt}
As shown in Fig.~\ref{fig:once_results}, when pre-trained by \method\ (solid lines), both SECOND~\cite{second} and CenterPoint~\cite{centerpoint} outperform training from scratch (dot lines) by considerable margins (2.70\% and 7.58\% mAP respectively). Meanwhile, increasing pre-training data also enlarges this gap, which again demonstrates the ability of \method\ to scale up.

\begin{table*}[t]
\vspace{0.25cm}
    \centering
        \caption{Few-shot performance on nuScenes validation set for \textbf{segmentation task} using 100\% pre-training data. We fine-tune on 5\% and 10\% nuScenes training data, respectively, and show the results of some of the categories.} 
        \vspace{-0.2cm}
        \label{tab:nuscseg_results}
        \centering
        \setlength\tabcolsep{8.0pt}
        \resizebox{0.99\linewidth}{!}
        {
            \begin{tabular}{c | c | c |c | c|  c| c| c| c| c}
                \toprule[1pt]
                Backbone & Method & Fine-tuning &mIOU & bus & car  & ped.  & trailer  &  sidewalk &  vegetable \\
                \cmidrule{1-10} 
                \multirow{8}{*}{Cylinder3D} 
                & From Scratch & 5\%  & 45.85 &  10.88 & 75.29  & 47.68 & 15.61 & 61.07 & 80.81\\     
                & BEV-MAE~\cite{bev-mae} & 5\% & 46.94 & 43.48 & 69.68 & 51.63 & 14.04 & 61.27 & 80.42\\
                & AD-PT~\cite{ad-pt} & 5\% & 45.61 & 9.33 & 76.08  & 51.27 & 15.95 & 60.49 & 79.67\\
                & SPOT (ours) & 5\% & \textbf{49.88} & \textbf{50.35} & \textbf{76.26}  & \textbf{52.42} & \textbf{16.45} & \textbf{63.74} & \textbf{81.83} \\ 
                \cmidrule{2-10}
                & From Scratch & 10\%             & 53.72 & 60.54 & 75.28   & 55.90 & 33.47 & 64.02 & 81.62\\     
                & BEV-MAE~\cite{bev-mae} & 10\%  & 53.75 & 57.11 & 76.26  & 54.88 & 20.92 & 65.00 & 81.81\\
                & AD-PT~\cite{ad-pt} & 10\%      & 52.86 & 53.76 & 81.09  & 53.11 & 28.60 & 65.45 & 82.14\\
                & SPOT (ours) &10\%  & \textbf{56.10} & \textbf{63.24} & \textbf{81.30}  & \textbf{57.86} & \textbf{33.99} & \textbf{67.04} & \textbf{82.73} \\ 
    
                \bottomrule[1pt]
            \end{tabular}
        }
\end{table*}

\vspace{-2pt}
\subsubsection{SemanticKITTI Segmentation}
\vspace{-1pt}
Results are presented in Table~\ref{tab:semtickitti_results}. It can be found that \method\ significantly improves mIoU metrics compared to training from scratch and achieves the best performance among all pre-training methods. For detailed categories, \method\ gains more than 20\% mIoU improvement compared to random initialization on truck, person and bicyclist, which can help guarantee safety in control task.

\subsubsection{nuScenes Segmentation} As shown in Table~\ref{tab:nuscseg_results}, considerable gains are achieved by \method, 4.03\% and 2.38\% mIOUs on $5\%$ and $10\%$ nuScenes data respectively. \method\ also achieves the best performance among all initialization methods.

\begin{table}[t]
    \small
	\centering	
    \setlength\tabcolsep{4pt}
        \caption{{Label-efficient pre-training setting, where SEMI and WS denote the semi-supervised and weakly-supervised pre-training setting, respectively. For downstream on nuScenes, only 5\% training data are used. L5\% denotes that we perform pre-training on 5\% sequence-level labeled data, while W5\% represents 5\% weakly-labeled data.}} 
        \vspace{-0.2cm}
        \label{tab:nusc_det_semi_results}
        \centering
        \resizebox{0.99\linewidth}{!}
        {
            \begin{tabular}{c | c | c | c | c | c }
                \toprule[1pt]
                Backbone & Method & P.D.A. & F.D.A. & mAP & NDS \\
                \cmidrule{1-6} 
                \multirow{8}{*}{SECOND~\cite{second}} 
                & From Scratch & - & 5\%  & 32.16 & 41.59 \\     
                & SPOT & L5\% & 5\%  & 37.96 & 48.45  \\    
                & SPOT & L20\% & 5\% & 39.63 & {51.63}  \\
                & SPOT & L100\% & 5\% & 42.57 & 54.28 \\ 
                & \textcolor{blue}{SPOT (SEMI)} & \textcolor{blue}{L5\% + W5\%} & \textcolor{blue}{5\%} & \textcolor{blue}{38.50} & \textcolor{blue}{50.03} \\ 
                & \textcolor{blue}{SPOT (SEMI)} & \textcolor{blue}{L5\% + W15\%} & \textcolor{blue}{5\%} & \textcolor{blue}{39.81} & \textcolor{blue}{51.51} \\ 

                & \textcolor{blue}{SPOT (SEMI)} &  \textcolor{blue}{L5\% + W95\%} & \textcolor{blue}{5\%} & \textcolor{blue}{42.18} & \textcolor{blue}{\textbf{54.44}} \\ 
                & \textcolor{blue}{SPOT (WS)} & \textcolor{blue}{W100\%} & \textcolor{blue}{5\%}  & \textcolor{blue}{\textbf{42.24}} & \textcolor{blue}{54.35}\\
                
                \cmidrule{1-6} 
                \multirow{8}{*}{CenterPoint~\cite{centerpoint}} 
                & From Scratch & - & 5\%  & 42.37 & 52.01 \\     
                & SPOT & L5\% & 5\%  & 43.56 & 53.04 \\    
                & SPOT & L20\% & 5\% & 44.94 & 54.95 \\ 
                & \textcolor{blue}{SPOT} &\textcolor{blue}{ L100\%} & \textcolor{blue}{5\%} & \textcolor{blue}{47.47} & \textcolor{blue}{57.11} \\ 
                & \textcolor{blue}{SPOT (SEMI)} & \textcolor{blue}{L5\% + W5\%} & \textcolor{blue}{5\%} & \textcolor{blue}{43.65} & \textcolor{blue}{53.82} \\ 
                & \textcolor{blue}{SPOT (SEMI)} & \textcolor{blue}{L5\% + W15\%} & \textcolor{blue}{5\%} & \textcolor{blue}{45.18} & \textcolor{blue}{54.98} \\ 
                & \textcolor{blue}{SPOT (SEMI)} &\textcolor{blue}{ L5\% + W95\%} & \textcolor{blue}{5\%} & \textcolor{blue}{\textbf{47.58}} & \textcolor{blue}{56.90} \\ 
                & \textcolor{blue}{SPOT (WS)} & \textcolor{blue}{W100\%} & \textcolor{blue}{5\%}  & \textcolor{blue}{47.56} & \textcolor{blue}{\textbf{57.18}}\\
                \bottomrule[1pt]
            \end{tabular}
        }
\end{table}

\begin{table}[t]
\vspace{0.25cm}
    \small
	\centering	
    \setlength\tabcolsep{6pt}
        \caption{Label-efficient pre-training setting. \textcolor{blue}{Fine-tuning process uses 20\% KITTI training data and we evaluate on KITTI validation set for detection task.}} 
        \vspace{-0.2cm}
        \label{tab:kitti_det_semi_results}
        \centering
        \resizebox{0.99\linewidth}{!}
        {
            \begin{tabular}{c | c | c | c | c }
                \toprule[1pt]
                Backbone & Method & P.D.A. & F.D.A. & mAP\\
                \cmidrule{1-5} 
                \multirow{8}{*}{SECOND~\cite{second}} 
                & From Scratch & - & 20\%  &  61.70  \\     
                & SPOT & L5\% & 20\%  & 63.53   \\    
                & SPOT & L20\% & 20\% & 65.45  \\ 
                & \textcolor{blue}{SPOT} &\textcolor{blue}{ L100\%} & \textcolor{blue}{20\%} & \textcolor{blue}{67.36} \\ 
                & SPOT (SEMI) & L5\% + W5\% & 20\% & 65.18  \\ 
                & SPOT (SEMI) & L5\% + W15\% & 20\% & {66.45}  \\ 
                & \textcolor{blue}{SPOT (SEMI)} &\textcolor{blue}{ L5\% + W95\%} & \textcolor{blue}{20\%} & \textcolor{blue}{67.66} \\ 
                & \textcolor{blue}{SPOT (WS)} & \textcolor{blue}{W100\%} & \textcolor{blue}{20\%}  & \textcolor{blue}{\textbf{67.68}} \\
                \cmidrule{1-5} 
                \multirow{8}{*}{PV-RCNN~\cite{pv-rcnn}} 
                & From Scratch & - & 20\%  & 66.71  \\     
                & SPOT & L5\% & 20\%  & 70.33  \\    
                & SPOT & L20\% & 20\% & 70.85  \\ 
                & \textcolor{blue}{SPOT} &\textcolor{blue}{ L100\%} & \textcolor{blue}{20\%} & \textcolor{blue}{71.77} \\ 
                & SPOT (SEMI) & L5\% + W5\% & 20\% & 70.40 \\ 
                & SPOT (SEMI) & L5\% + W15\% & 20\% & {70.86} \\ 
                & \textcolor{blue}{SPOT (SEMI)} &\textcolor{blue}{ L5\% + W95\%} & \textcolor{blue}{20\%} & \textcolor{blue}{71.67} \\ 
                & \textcolor{blue}{SPOT (WS)} & \textcolor{blue}{W100\%} & \textcolor{blue}{20\%}  & \textcolor{blue}{\textbf{72.00}} \\
                \bottomrule[1pt]
            \end{tabular}
        }
\end{table}

\begin{table}[h]
    \small
	\centering	
    \setlength\tabcolsep{6pt}
        \caption{Label-efficient pre-training setting. \textcolor{blue}{Fine-tuning process uses 5\% nuScenes training data and we evaluate on nuScenes validation set for segmentation downstream task.}} 
        \vspace{-0.2cm}
        \label{tab:nusc_seg_semi_results}
        \centering
        \resizebox{0.99\linewidth}{!}
        {
            \begin{tabular}{c | c | c | c | c }
                \toprule[1pt]
                Backbone & Method & P.D.A.& F.D.A. & mIOU \\
                \cmidrule{1-5} 
                \multirow{8}{*}{Cylinder3D~\cite{cylinder3d}} 
                & From Scratch & - & 5\%  & 45.85 \\     
                & SPOT & L5\% & 5\%  & 46.71 \\    
                & SPOT & L20\% & 5\% & 47.84 \\ 
                & \textcolor{blue}{SPOT} &\textcolor{blue}{ L100\%} & \textcolor{blue}{5\%} & \textcolor{blue}{49.88} \\ 
                & SPOT (SEMI) & L5\% + W5\% & 5\% & 47.60 \\ 
                & SPOT (SEMI) & L5\% + W15\% & 5\% & {48.84} \\
                & \textcolor{blue}{SPOT (SEMI)} &\textcolor{blue}{ L5\% + W95\%} & \textcolor{blue}{5\%} & \textcolor{blue}{50.17} \\ 
                & \textcolor{blue}{SPOT (WS)} & \textcolor{blue}{W100\%} & \textcolor{blue}{5\%}  & \textcolor{blue}{\textbf{51.07}} \\
                \bottomrule[1pt]
            \end{tabular}
        }
\end{table}

\begin{table}[tb!]
    \small
	\centering	
    \setlength\tabcolsep{7pt}
        \caption{Label-efficient pre-training setting. \textcolor{blue}{Fine-tuning process uses 10\% SemanticKITTI training data and we evaluate on SemanticKITTI validation set for segmentation downstream task.}}
        \vspace{-0.2cm}
        \label{tab:semtickitti_semi_results}
        \centering
        \resizebox{0.99\linewidth}{!}
        {
            \begin{tabular}{c | c | c | c | c }
                \toprule[1pt]
                Backbone & Method & P.D.A.& F.D.A. & mIOU \\
                \cmidrule{1-5} 
                \multirow{8}{*}{Cylinder3D~\cite{cylinder3d}} 
                & From Scratch & - & 10\%  & 49.01 \\     
                & SPOT & L5\% & 10\%  & 52.50 \\    
                & SPOT & L20\% & 10\% & 54.10 \\ 
                & \textcolor{blue}{SPOT} &\textcolor{blue}{ L100\%} & \textcolor{blue}{10\%} & \textcolor{blue}{55.58} \\ 
                & SPOT (SEMI) & L5\% + W5\% & 10\% & 53.62 \\ 
                & SPOT (SEMI) & L5\% + W15\% & 10\% & {54.70} \\
                & \textcolor{blue}{SPOT (SEMI)} &\textcolor{blue}{ L5\% + W95\%} & \textcolor{blue}{10\%} & \textcolor{blue}{55.96} \\ 
                & \textcolor{blue}{SPOT (WS)} & \textcolor{blue}{W100\%} & \textcolor{blue}{10\%}  & \textcolor{blue}{\textbf{56.18}} \\
                \bottomrule[1pt]
            \end{tabular}
        }
\end{table}

\subsection{From Supervised to Semi-supervised \textcolor{red}{and Weakly-supervised} Pre-training}
\label{exp:semi_exp}

In this section, considering that SPOT requires supervised information (for the purpose of generating dense occupancy) to perform the 3D pre-training task, we study SPOT's dependence on pre-training supervision information. In order to demonstrate \method's ability to scale up, we design experiments to explore \textcolor{blue}{the semi-supervised and weakly-supervised setting during the pre-training phase.}  

\textbf{For semi-supervised pre-training setting}, we first pre-train the backbone with SPOT using only 5\% sequence-level labeled data and \textcolor{blue}{5\%, 15\%, 95\%} sequence-level unlabeled data, where the unlabeled data are pseudo-labeled~\cite{pseudo-label} by employing a naive mean-teacher approach~\cite{mean-teacher}. Refer to~\cite{mean-teacher} for more details in calculating the pseudo-labels. 

\textcolor{red}{\textbf{For weakly-supervised pre-training setting}, we first employ vision foundation models~\cite{ravi2024sam,liu2025grounding} to obtain semantic labels for foreground and background objects of the 2D image. Then we establish correspondences between the 2D image space and 3D point space to transfer 2D semantic labels to 3D point cloud data. In our practical implementation, we utilize the 3D projection API function\footnote{https://github.com/waymo-research/waymo-open-dataset} provided by the Waymo dataset~\cite{waymo}, which compensates for point cloud motion, alongside effectively correcting distortion and deformation caused by the camera rolling shutter effect, thereby ensuring precise registration between camera and LiDAR data. Through this projection mapping relationship, we associate 3D point cloud data with semantic labels from the image plane, subsequently generating the occupancy labels required for SPOT pre-training.
}

After the pre-training phase, the pre-trained backbone is fine-tuned on downstream tasks including nuScenes, KITTI  detection tasks, and nuScenes and SemanticKITTI segmentation tasks using different baseline models. 

The experimental results of semi-supervised \textcolor{blue}{and weakly-supervised} pre-training setting are reported in Tables~\ref{tab:nusc_det_semi_results},~\ref{tab:kitti_det_semi_results},~\ref{tab:nusc_seg_semi_results}, and~\ref{tab:semtickitti_semi_results}. It can be found that semi-supervised \textcolor{blue}{and weakly-supervised} pre-training with \method\ achieves comparable downstream performance as that of fully-supervised pre-training. It consistently improves different architectures on various datasets and tasks. Also, when incorporating more weakly-labeled data to perform the pre-training (\textit{e.g.}, comparing L5\%+W5\%, L5\%+W15\% and \textcolor{blue}{L5\%+W95\%}), the performance of the downstream task significantly improves. \textcolor{blue}{Notably, weakly-supervised pre-training (W100\%) also demonstrates competitive performance without using any human-annotated labels during the pre-training phase.} Thus, we believe that \method\ is able to generalize to label-efficient pre-training settings and further attain performance scalability on different downstream datasets and tasks such as 3D detection and segmentation tasks.

Overall, we conclude that SPOT requires a certain amount of supervised information in the pre-training dataset, but it remains compatible with unlabeled data (as observed in Table~\ref{tab:nusc_det_semi_results} to~\ref{tab:semtickitti_semi_results}). From another perspective, SPOT can alleviate the model's reliance on human annotations in downstream tasks while achieving better model performance, thereby reducing the annotation costs associated with these downstream tasks.

\begin{table*}[t]
    \caption{{\color{blue}{Fine-tuing on 100\% KITTI training data and evaluating on KITTI validation set with 40 recall positions at moderate difficulty level. SPOT (WS) means we pre-train SPOT without manually labeled data, using generated labels by the completely weakly-supervised method as described in Section~\ref{exp:semi_exp}. UN means unsupervised method.}}}
    \label{tab:kitti_occmae}
    \vspace{-0.2cm}
    \setlength\tabcolsep{16.0pt}
    \centering
    \resizebox{0.98\linewidth}{!}
    {\color{blue}{{
        \begin{tabular}{c | c | c | c | c | c | c }
            \toprule[1pt]
             {Backbone} & {Method} & {F.D.A.} & mAP & {Car} & {Pedestrian} & {Cyclist} \\
            \cmidrule{1-7}
            \multirow{3}{*}{SECOND~\cite{second}}  & From Scratch & 100\% & 65.35 & 81.50  & 48.82 & 65.72 \\
            & Occupancy-MAE (UN)~\cite{occ-mae} & 100\% & 68.24 & 81.98 & 53.67  & 69.08  \\
            & SPOT (WS) & 100\% & \textbf{69.73} & \textbf{82.52} & \textbf{56.72} & \textbf{69.95}  \\

            \cmidrule{1-7}
            \multirow{3}{*}{PV-RCNN~\cite{pv-rcnn}} & From Scratch & 100\% & 70.57  & 84.50 & 57.06 & 70.14 \\
            
            & Occupancy-MAE (UN)~\cite{occ-mae} & 100\% & \textbf{73.29}  & 84.82 & 59.07  & \textbf{75.68}  \\

            & SPOT (WS) & 100\% & 73.14  & \textbf{84.86}  & \textbf{61.29} &  73.27  \\
           \bottomrule[1pt]
        \end{tabular}
    }}}
\end{table*}

\begin{table*}[t]
    \caption{{\color{blue}{Fine-tuing on 100\% KITTI training data and evaluating on KITTI validation set with AP calculated by 11 recall positions evaluating bounding box and orientation.}}} 
    \vspace{-0.2cm}
    \label{tab:kitti_bbox_aos}
    \centering
    \setlength\tabcolsep{6.0pt}
    \resizebox{1.0\linewidth}{!}
    {\color{blue}{{
        \begin{tabular}{c | c  | c | c c c | c c c | c c c }
            \toprule[1pt]
             \multirow{2}{*}{Evaluation} & \multirow{2}{*}{Method} & \multirow{2}{*}{F.D.A.}  & \multicolumn{3}{c|}{Car} & \multicolumn{3}{c|}{Pedestrian} & \multicolumn{3}{c}{Cyclist} \\
            &  & & Easy & Mod. & Hard & Easy & Mod. & Hard & Easy & Mod. & Hard \\
            \cmidrule{1-12}
            \multirow{3}{*}{bbox}  & SECOND~\cite{second} & 100\% & 90.73 &  89.76  & 88.94 & 68.70 &  65.27 & 62.52  & 87.88  & 75.43  & 71.67\\
            & Occupancy-MAE (UN)~\cite{occ-mae} + SECOND & 100\% & 94.81 &  89.98 & 89.35 & 70.37 & 67.45 & 65.14 & 91.82 & \textbf{78.65} & 73.77 \\
            & SPOT (WS) + SECOND & 100\% & \textbf{95.48} & \textbf{90.22} & \textbf{89.43} & \textbf{72.94} &  \textbf{69.08} & \textbf{66.31} & \textbf{93.64} & {77.34} & \textbf{74.03} \\
             \cmidrule{1-12}
            \multirow{3}{*}{aos}  & SECOND~\cite{second} & 100\% & 90.73  & 89.63 &  88.70  & 63.46 &  60.13 &  56.93  & 87.63  & 74.67  & 71.00\\
            & Occupancy-MAE (UN)~\cite{occ-mae} + SECOND & 100\% & 94.66 & 89.88 & 88.92 & 65.33 & 61.55 & 59.23 & 91.57 & \textbf{78.42} & 73.50 \\
            & SPOT (WS) + SECOND & 100\% & \textbf{95.44} & \textbf{90.14} & \textbf{89.26} & \textbf{70.03} & \textbf{65.45} & \textbf{62.13} & \textbf{93.19} & {76.79} & \textbf{73.51} \\

           \bottomrule[1pt]
        \end{tabular}
    }}}

\end{table*}

\begin{table}[t]
    \centering
        \caption{{\color{blue}{Fine-tuing on 100\% nuScenes detection training data and evaluating on nuScenes detection validation set.}}}
        \vspace{-0.2cm}
        \label{tab:nusc_occmae}
        \centering
        \resizebox{0.99\linewidth}{!}
        {\color{blue}{{{
            \begin{tabular}{c |  c | c | c | c }
                \toprule[1pt]
                Backbone & Method &  {F.D.A.} & mAP & NDS \\
                \cmidrule{1-5} 
                \multirow{3}{*}{CenterPoint~\cite{centerpoint}} 
                & From Scratch & 100\% & 56.0 & 64.5 \\     
                & Occupancy-MAE (UN)~\cite{occ-mae} & 100\% & 56.5 & 65.0 \\    
                & SPOT (WS) & 100\% & \textbf{57.5} & \textbf{65.3}  \\   
                \bottomrule[1pt]
            \end{tabular}
        }}}}
\end{table}

\begin{table}[t]
    \centering
        \caption{\color{blue}{Fine-tuing on 100\% nuScenes segmentation training data and evaluating on nuScenes segmentation validation set.}}
        \vspace{-0.2cm}
        \label{tab:nusc_seg_occmae}
        \centering
        \resizebox{0.99\linewidth}{!}
        {\color{blue}{{{
            \begin{tabular}{c | c | c | c | c  }
                \toprule[1pt]
                Backbone & Method  & F.D.A & Epoch & mIOU \\
                \cmidrule{1-5} 
                \multirow{6}{*}{Cylinder3D~\cite{cylinder3d}} 
                & From Scratch  & 100\% & 15 & 70.22 \\       
                & Occupancy-MAE (UN)~\cite{occ-mae} & 100\% & 15 & 71.61 \\   
                & SPOT (WS) & 100\% & 15 & \textbf{72.37}  \\   
                \cmidrule{2-5}
                & From Scratch  & 100\% & 25 & 70.83 \\   
                & Occupancy-MAE (UN)~\cite{occ-mae} & 100\% & 25 &  72.85 \\  
                & SPOT (WS) & 100\% & 25 & \textbf{73.26} \\   
 
                \bottomrule[1pt]
            \end{tabular}
        }}}}
\end{table}

{\color{blue}{
\subsection{Comparison with Occupancy-based Pre-training Methods}
\label{exp:compare_occmae}

To validate the effectiveness of our pre-training strategy compared to binary occupancy prediction methods, we conduct extensive experiments on both detection and segmentation downstream tasks. Specifically, we evaluate our method \textbf{SPOT (WS)}, which is pre-trained using the auto-generated labels without manually labeled data (see Sec.~\ref{exp:semi_exp} for details on SPOT's unlabeled pre-training), against the recent binary occupancy prediction pre-training method, Occupancy-MAE~\cite{occ-mae}. All the experimental settings in this section are consistent with those reported in their paper.

\subsubsection{Detection Results}
As shown in Tables~\ref{tab:kitti_occmae} and~\ref{tab:kitti_bbox_aos}, we evaluate the detection performance on the KITTI validation set using SECOND and PV-RCNN as detectors. For SECOND, our method achieves 69.73\% mAP on moderate difficulty, outperforming Occupancy-MAE~\cite{occ-mae} by 1.49\% (69.73\% vs 68.24\%). The improvements are consistent across different categories, with notable gains of 0.54\% on Car, 3.05\% on Pedestrian, and 0.87\% on Cyclist. Similar trends can be observed on PV-RCNN, where our method obtains better performance on Car and Pedestrian categories while maintaining competitive results on Cyclist detection.

We further validate the effectiveness on the more challenging nuScenes dataset. As shown in Table~\ref{tab:nusc_occmae}, with CenterPoint as the detector, our method achieves 57.5\% mAP and 65.3\% NDS, surpassing Occupancy-MAE~\cite{occ-mae} by 1.0\% and 0.3\% respectively. The consistent improvements across different datasets and backbone networks demonstrate that our semantic occupancy prediction provides richer supervision signals than binary occupancy prediction for learning transferable representations.

\subsubsection{Segmentation Results} 
To evaluate the generalization ability of our pre-training strategy, we also conduct experiments on the segmentation task using Cylinder3D as the backbone. As shown in Table~\ref{tab:nusc_seg_occmae}, our method consistently outperforms Occupancy-MAE~\cite{occ-mae} under different training epochs. Specifically, with 15 epochs of fine-tuning, our method achieves 72.37\% mIOU, surpassing Occupancy-MAE~\cite{occ-mae} by 0.76\%. When training for longer epochs, the performance gap is maintained (73.26\% vs 72.85\%). The superior segmentation results further verify that learning to predict semantic occupancy helps the model better understand the 3D scene structure and semantic information, which benefits various downstream tasks.

The consistent improvements compared with Occupancy-MAE~\cite{occ-mae} across different tasks, backbones and datasets demonstrate the superiority of our SPOT pre-training framework. Unlike Occupancy-MAE that requires dataset-dependent pre-training before fine-tuning on each specific downstream task, our framework advocates a unified "one-to-many" paradigm where a single pre-training on Waymo dataset enables effective transfer to various downstream tasks and datasets. More importantly, even under the same label-free setting, SPOT consistently outperforms Occupancy-MAE by significant margins across multiple tasks (detection, segmentation) and datasets (KITTI, nuScenes), which underscores the effectiveness of our SPOT framework in leveraging semantic-aware occupancy prediction to learn robust, transferable, and domain-agnostic 3D representations.
}}


\begin{table*}[t]
\vspace{0.25cm}
    \small
	\centering	
    \setlength\tabcolsep{6pt}
        \caption{The impact of pre-training task superiority, where we employ the detection pre-training, segmentation pre-training, occupancy pre-training, respectively. We perform fine-tuning experiments on multiple datasets of both detection and segmentation tasks, using 100\% pre-training data.} 
        \vspace{-0.15cm}
        \label{tab:abl_task}
        \resizebox{0.90\linewidth}{!}
        {
        \begin{tabular}{c | c | c  c | c | c}
            \toprule[1pt]
            \multirow{2}{*}{Different Pre-training Tasks}  & KITTI (det) & \multicolumn{2}{c|}{nuScenes (det)}  & SemanticKITTI (seg) & nuScenes (seg)\\
            \cmidrule(l){2-6}
             & mAP (mod.) &  mAP & NDS & mIoU & mIoU \\
            \cmidrule{1-6}
            Without Pre-training & 61.70 & 42.37 & 52.01 & 60.60 & 69.15\\
            Detection Pre-training & 65.46 & 40.89 & 49.75 & 60.20 & 69.31\\
            Segmentation Pre-training & 58.13 & 36.23 & 47.01 & 61.95 & 69.60\\
            Occupancy Pre-training & \textbf{67.36} & \textbf{47.47} & \textbf{57.11} & \textbf{62.24} & \textbf{70.77}  \\ 
            \bottomrule[1pt]
        \end{tabular}
        }
\end{table*}

\begin{table*}[t]
    \small
	\centering	
    \setlength\tabcolsep{4pt}
    \begin{small}
        \caption{Ablation study on pre-training strategies across different datasets.}
        \vspace{-0.15cm}
        \label{tab:abl_results}
        \resizebox{0.92\linewidth}{!}
        {
        \begin{tabular}{c| c| c| c | c  c | c | c}
            \toprule[1pt]
            \multirow{2}{*}{Occupancy Prediction} & \multirow{2}{*}{Loss Balancing}   & \multirow{2}{*}{Beam Re-sampling} & \multirow{2}{*}{Dataset Balancing} & \multicolumn{2}{c|}{nuScenes} & \multicolumn{1}{c|}{ONCE} & \multicolumn{1}{c}{KITTI}\\
            \cmidrule(l){5-8}
              &   &  & &  mAP & NDS & mAP & mAP (mod.) \\
            \cmidrule{1-8}
             &  &   &    & 32.16  & 41.59  & 35.96 &  61.70 \\ 
             \checkmark  &  &    & & 36.55 & 46.98 & 36.00 & 63.70\\
             \checkmark & \checkmark &   &  & 37.90  & 47.82  & 37.30 &  64.70 \\
              \checkmark & \checkmark &  \checkmark &  & 38.63  & 48.85  & 39.19 &  65.92 \\
               \checkmark & \checkmark &  \checkmark & \checkmark& \textbf{40.39}  &  \textbf{51.65}  &  \textbf{40.63}  &  \textbf{66.45}  \\
            \bottomrule[1pt]
        \end{tabular}
        }
    \end{small}
\end{table*}

\begin{table}[t]
    \centering
    \begin{small}
        \caption{{Experiments of extending the training schedule on nuScenes for detection task.}}
        \vspace{-0.15cm}
        \label{tab:train_schedule}
        \resizebox{0.99\linewidth}{!}
        {
        \begin{tabular}{c | c | c | c | c | c}
            \toprule[1pt]
             Detector  & Method & P.D.A. & Training Schedule & mAP & NDS \\
            \cmidrule{1-6}
            \multirow{4}{*}{SECOND~\cite{second}} 
             &  From Scratch & - & 30 epochs & 32.16 & 41.59\\
              &  From Scratch & - & 150 epochs & 36.79 & 51.01\\
             &  SPOT (ours) & 20\% & 30 epochs & 39.63 & 51.63\\ 
             &  SPOT (ours) & 100\% & 30 epochs & \textbf{42.57} & \textbf{54.28}\\ 
            \cmidrule{1-6}
            \multirow{4}{*}{CenterPoint~\cite{centerpoint}} 
             & From Scratch &  -  & 30 epochs & 42.37 & 52.01  \\
              & From Scratch &  -  & 150 epochs & 41.01 & 53.92 \\ 
              & SPOT (ours) &  20\% & 30 epochs & 44.94 & 54.95  \\ 
             & SPOT (ours) &  100\% & 30 epochs & \textbf{47.47} &  \textbf{57.11}   \\ 
            \cmidrule{1-6}
            \multirow{3}{*}{DSVT~\cite{wang2023dsvt}} 
             & From Scratch &  -  & 20 epochs & 49.78 & 58.63 \\ 
              & From Scratch &  -  & 150 epochs & 54.30 & \textbf{63.58} \\ 
              & SPOT (ours) &  20\% & 20 epochs & \textbf{56.65} &  63.52  \\ 
            \bottomrule[1pt]
        \end{tabular}
        }
    \end{small}
\end{table}

\subsection{Discussions and Analyses}
\label{exp:discussion}

\subsubsection{Discussions of Pre-training} 

\noindent \textbf{Pre-training by Different Tasks.} We argue that occupancy prediction is a scalable and general task for 3D representation learning. Here we conduct experiments to compare different kinds of existing task for pre-training, including detection and segmentation tasks. Pre-training is conducted on the full Waymo dataset. Besides, fine-tuning setting employs $20\%$ KITTI data, $5\%$ nuScenes(det) data, $100\%$ SemanticKITTI data, and $100\%$ nuScenes(seg) data. The results presented in Table~\ref{tab:abl_task} reveal that relying solely on detection as a pre-training task yields minimal performance gains, particularly when significant domain discrepancies exist, \textit{e.g.} Waymo to nuScenes. Similarly, segmentation alone as a pre-training task demonstrates poor performance in the downstream detection task, likely due to the absence of localization information. On the contrary, our occupancy prediction task is beneficial to achieve consistent performance improvements for various datasets and tasks.

\noindent \textbf{Fine-tuning Experiments on Extending the Training Schedule.} To further demonstrate that our pre-training method enhances the backbone capacity rather than simply accelerating the convergence speed of training model, we consider conducting experiments under different training schedules. We select SECOND~\cite{second}, CenterPoint~\cite{centerpoint}, and DSVT~\cite{wang2023dsvt}, as the baseline method, and the experimental results are shown in Table~\ref{tab:train_schedule}. It can be seen from these results that, the results of only training 30 epochs using our SPOT pre-training can exceed the results of 150 epochs of training from scratch by $2.35\% \sim 5.78\%$.


\begin{table}[t!]
    \centering
    \setlength\tabcolsep{12pt}
        \caption{{Pre-training on nuScenes and fine-tuning on KITTI for detection. We fine-tune on 20\% training data.}} 
        \vspace{-0.15cm}
        \label{tab:kitti_nuscocc_results}
        \centering
        \resizebox{0.92\linewidth}{!}
        {
            \begin{tabular}{c | c | c | c}
                \toprule[1pt]
                Backbone & Method & F.D.A. & mAP \\
                \cmidrule{1-4} 
                \multirow{2}{*}{SECOND~\cite{second}} 
                & From Scratch &  20\%  & 61.70  \\     
                & SPOT (ours)  & 20\% & \textbf{64.39} \\    
                \cmidrule{1-4} 
                \multirow{2}{*}{PV-RCNN~\cite{pv-rcnn}} 
                & From Scratch & 20\%  & 66.71  \\     
                & SPOT (ours)  & 20\% & \textbf{69.58} \\    
                \bottomrule[1pt]
            \end{tabular}
        }
\end{table}

\begin{table}[t]
    \centering
        \caption{{Fine-tuning performance on nuScenes benchmark for detection task based on the binary occupancy pre-training. We fine-tune on 5\% training data.}} 
        \vspace{-0.2cm}
        \label{tab:nusc_det_01occ_results}
        \centering
        \resizebox{0.99\linewidth}{!}
        {
            \begin{tabular}{c | c | c | c | c }
                \toprule[1pt]
                Backbone & Method & F.D.A. & mAP & NDS \\
                \cmidrule{1-5} 
                \multirow{3}{*}{CenterPoint~\cite{centerpoint}} 
                & From Scratch & 5\%  & 42.37 & 52.01 \\     
                & Binary Pre-training & 5\% & 42.05 & 51.63 \\    
                & SPOT (ours) & 5\% & \textbf{44.94} & \textbf{54.95}  \\   
                \bottomrule[1pt]
            \end{tabular}
        }
\end{table}

\begin{table}[tb!]
    \centering
    \setlength\tabcolsep{6pt}
    \begin{small}
        \caption{{Fine-tuning performance of employing transformer-based structure on different datasets.}}
        \vspace{-0.15cm}
        \label{tab:DSVT_results}
        \resizebox{0.99\linewidth}{!}
        {
        \begin{tabular}{c | c | c | c | c | c}
            \toprule[1pt]
            \multirow{2}{*}{Detector} & \multirow{2}{*}{Method} &\multirow{2}{*}{P.D.A.} & \multicolumn{2}{c|}{nuScenes} & \multicolumn{1}{c}{ONCE}\\
            \cmidrule(l){4-6}
               &  & &  mAP & NDS & mAP\\
            \cmidrule{1-6}
            \multirow{3}{*}{DSVT~\cite{wang2023dsvt}} 
              &  From Scratch & - & 49.78 & 58.63  & 51.52 \\ 
              & SPOT (ours) &  5\%  & 55.47 & 62.17  & 57.81 \\ 
              & SPOT (ours) &  20\%  & \textbf{56.65} &\textbf{ 63.52} & \textbf{59.78} \\ 
            \bottomrule[1pt]
        \end{tabular}
        }
    \end{small}
\end{table}

\begin{table*}[t]
\vspace{0.25cm}
    \centering
    \setlength\tabcolsep{10pt}
    \caption{Fine-tuning performance on Waymo benchmark (LEVEL\_2 metric). We fine-tune on 3\% Waymo training data. P.D.A. represents the Pre-training Data Amount.}
    \vspace{-0.15cm}
    \label{tab:waymo_results}
    \begin{small}
    \resizebox{0.99\linewidth}{!}
    {
        \begin{tabular}{c | c | c | c | c c c}
            \toprule[1pt]
            \multirow{2}{*}{Backbone} & \multirow{2}{*}{Method} & \multirow{2}{*}{P.D.A.} & \multicolumn{4}{c}{L2 AP / APH}  \\
            \cmidrule(l){4-7}
            & & & Overall & Vehicle & Pedestrian & Cyclist \\
            \cmidrule{1-7}
            \multirow{6}{*}{CenterPoint~\cite{centerpoint}}   & From Scratch    & - & 59.00 / 56.29 & 57.12 / 56.57& 58.66 / 52.44 & 61.24 / 59.89 \\
              & BEV-MAE~\cite{bev-mae} & 100\% & 59.51 / 56.81 & 57.38 / 56.84 & 58.87 / 52.78 & 62.28 / 60.82 \\
              & AD-PT~\cite{ad-pt} & 100\% & 61.21 / 58.46 & 60.35 / 59.79 & 60.57 / 54.02 & 62.73 / 61.57 \\
              & SPOT (ours) & 5\% & 61.61 / 58.69 & 58.63  / 58.06 & 61.35 / 54.53 & 64.86 / 63.48 \\
              & SPOT (ours) & 20\% & 62.74 / 59.84 & 59.67 / 59.09 & 62.73 / 56.01 & 65.83 / 64.41 \\
              &  SPOT (ours) & 100\% & \textbf{63.76 / 60.98} & \textbf{61.17 / 60.63} &  \textbf{64.05 / 57.49} & \textbf{66.07 / 64.81} \\
            \bottomrule[1pt]
        \end{tabular}
    }
    \end{small}
\end{table*}

\begin{figure*}[tb!]
    \centering
    \small
    \subfloat[Fine-tuning results on nuScenes using mAP metric.]{
    \begin{minipage}{0.41\linewidth}{\begin{center}
    \resizebox{\linewidth}{!}{\includegraphics{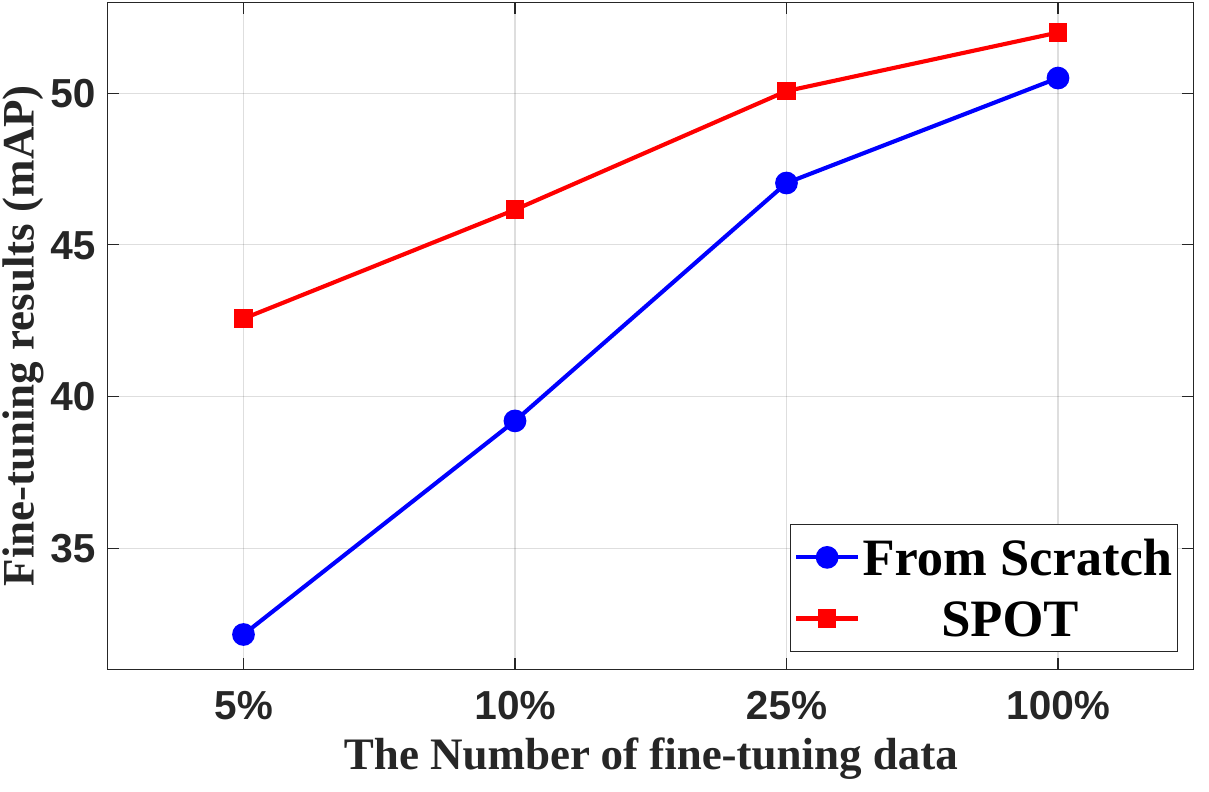}}\end{center}}\end{minipage}
    }
    \subfloat[Fine-tuning results on nuScenes using NDS metric]{
    \begin{minipage}{0.41\linewidth}{\begin{center}
    \resizebox{\linewidth}{!}{\includegraphics{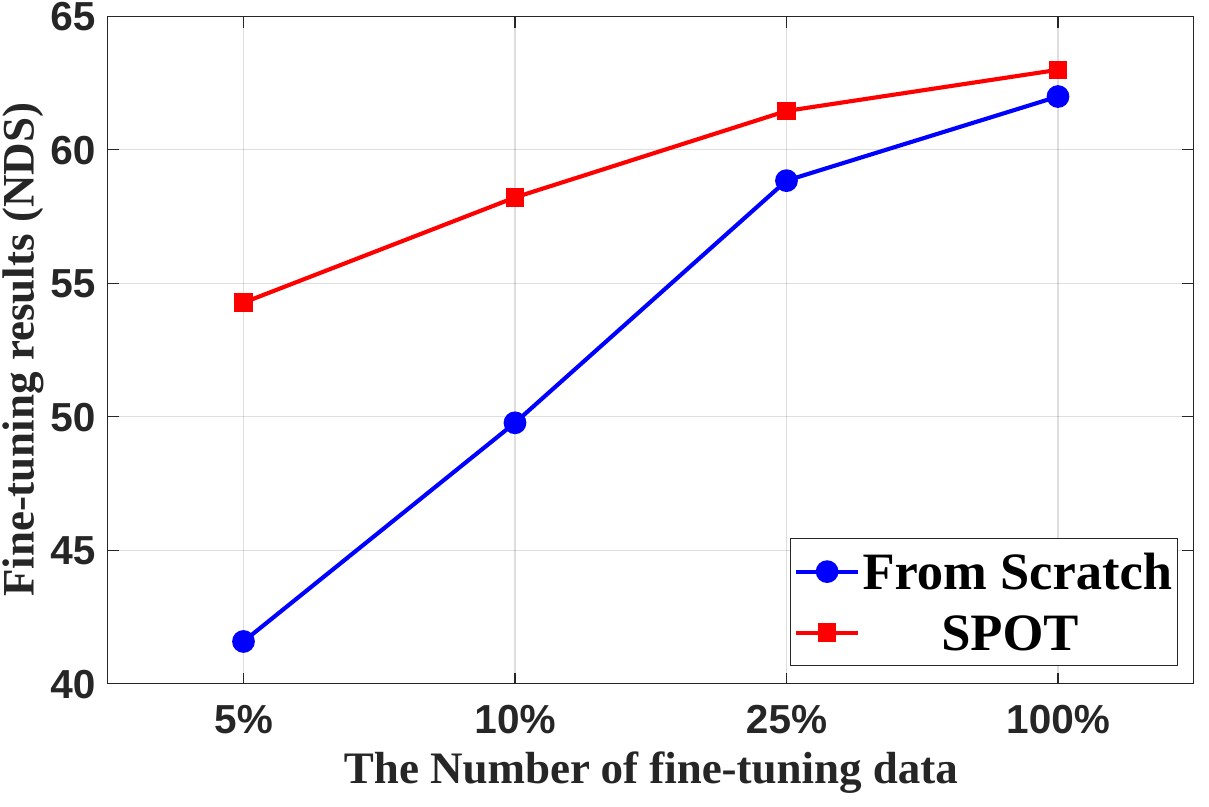}}\end{center}}\end{minipage}
    }
    \vspace{-0.10cm}
    \caption{Fine-tuning performance on nuScenes dataset for detection task with different numbers of annotated data.}
    \label{fig:nusc_scale_results}
\end{figure*}

\noindent \textbf{Pre-training on nuScenes Dataset.}
To verify that SPOT is able to pre-train on other datasets, we utilize the model that is pre-trained on Waymo to predict occupancy labels on nuScenes dataset and generate pseudo occupancy labels. Next, we pre-train SPOT from scratch on such nuScenes data, and then fine-tune on the 20\% KITTI data. As shown in Table~\ref{tab:kitti_nuscocc_results}, SPOT achieves significant gains compared to baseline results on KITTI dataset, demonstrating the effectiveness and generalization of SPOT.

\noindent \textbf{Pre-training with Binary Occupancy Labels.}
In this part, we conduct additional experiments using 20\% sequence-level binary occupancy-based Waymo data to perform the pre-training, and employ 5\% nuScenes data for downstream fine-tuning. For consistency with previous experiments, we use the widely-adopted CenterPoint detector~\cite{centerpoint} as our baseline. The results are shown in  Table~\ref{tab:nusc_det_01occ_results}. It can be seen that, simple binary occupancy prediction does not bring performance gains when it performs cross-domain experiments, such as Waymo to nuScenes. This is mainly due to that, the pre-training model is difficult to learn semantically-rich information of the 3D scene when only employing the binary occupancy prediction as pre-training task. These findings highlight the importance of carefully considering and optimizing the pre-training process to achieve superior performance in the subsequent tasks.

\begin{table}[tb!]
\vspace{0.20cm}
    \small
	\centering	
    \setlength\tabcolsep{12pt}
        \caption{Fine-tuning performance on KITTI and nuScenes (det) benchmark with 100\% data, using SECOND.} 
        \vspace{-0.15cm}
        \label{tab:det_100_results}
        \centering
        \resizebox{0.92\linewidth}{!}
        {
            \begin{tabular}{c | c | c c }
                \toprule[1pt]
                \multirow{2}{*}{Method}  & \multicolumn{1}{c|}{KITTI} & \multicolumn{2}{c}{nuScenes (det)}\\
                \cmidrule(l){2-4}
                & mAP(mod.) & mAP & NDS\\
                \cmidrule{1-4} 
                From Scratch & 66.70 & 50.59 & 62.29 \\ 
                SPOT (ours) & \textbf{68.57} & \textbf{51.88} & \textbf{62.68} \\ 
                \bottomrule[1pt]
            \end{tabular}
        }
\end{table}

 \begin{table}[tb!]
    \small
	\centering	
    \setlength\tabcolsep{12pt}
        \caption{Fine-tuning performance on SemanticKITTI and nuScenes (seg) benchmark with 100\% data.} 
        \vspace{-0.15cm}
        \label{tab:seg_100_results}
        \centering
        \resizebox{0.92\linewidth}{!}
        {
            \begin{tabular}{c | c | c }
                \toprule[1pt]
                \multirow{2}{*}{Method}  & \multicolumn{1}{c|}{SemanticKITTI} & \multicolumn{1}{c}{nuScenes (seg)}\\
                \cmidrule(l){2-3}
                & mIOU & mIOU\\
                \cmidrule{1-3} 
                From Scratch & 60.60 & 69.15 \\ 
                SPOT (ours) & \textbf{62.24} & \textbf{70.77} \\ 
                \bottomrule[1pt]
            \end{tabular}
        }
\end{table}

\begin{figure*}[h]
    \centering
    \small
    \resizebox{0.96\linewidth}{!}
    {\includegraphics{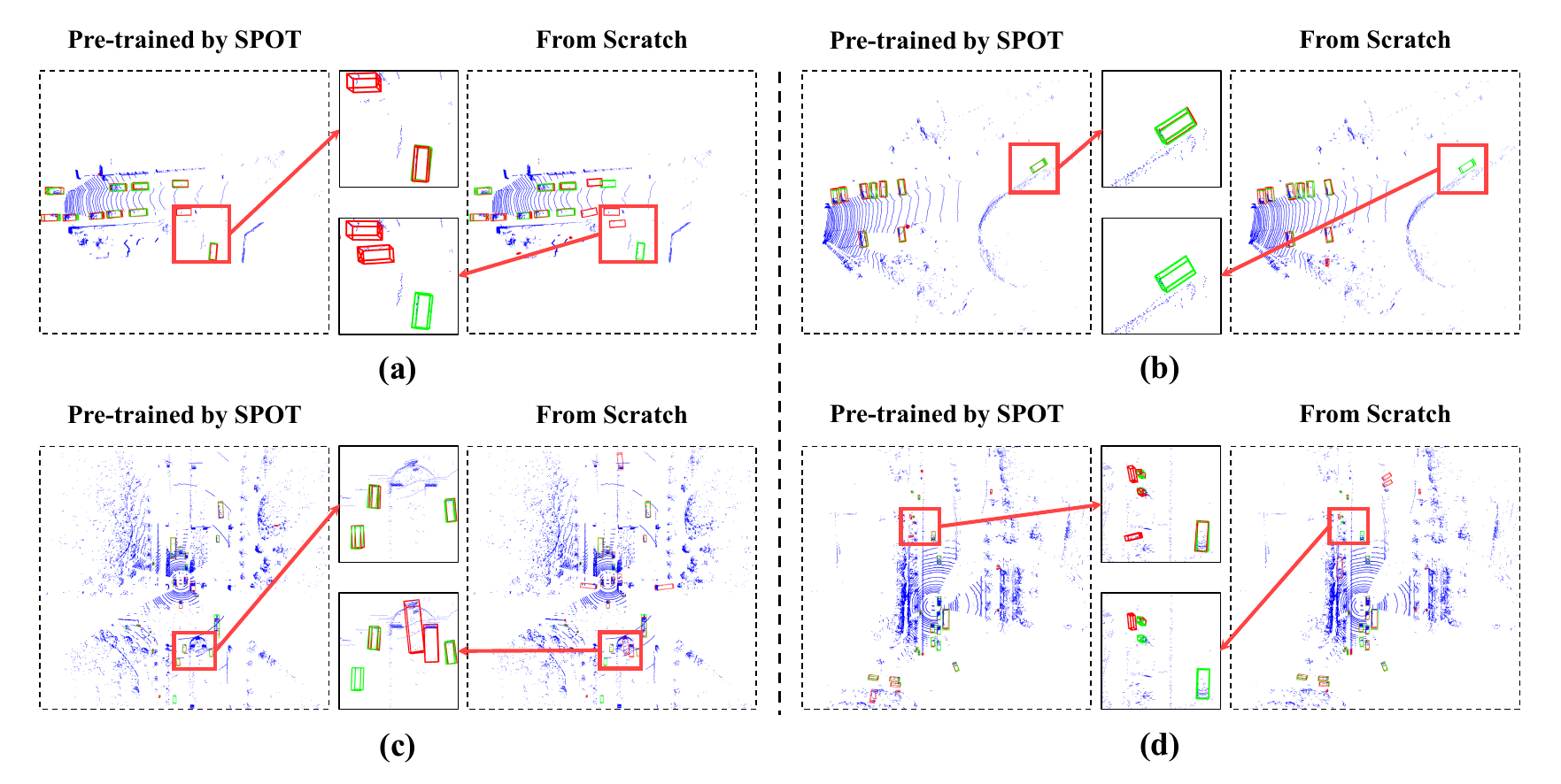}}
    \vspace{-0.25cm}
    \caption{Visualization of downstream detection results, where the red and green boxes correspond to the predicted results and the ground truth, respectively. (a) and (b) are the results of KITTI, (c) and (d) are the results of ONCE.}
    \label{fig:det_vis}
\end{figure*}

\begin{figure}[h]
    \centering
    \small
    \resizebox{0.98\linewidth}{!}
    {\includegraphics{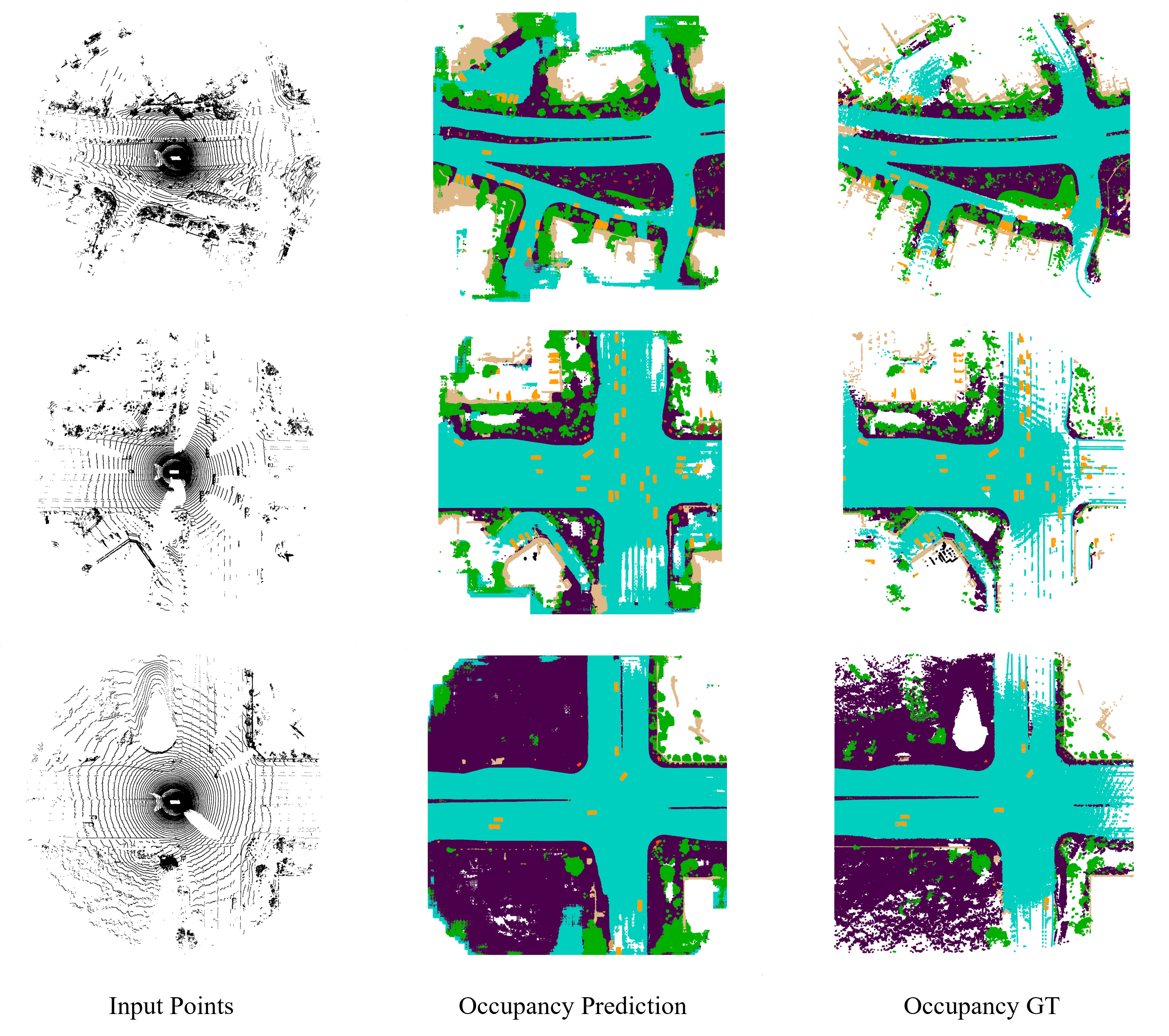}}
    \caption{Visualizing occupancy prediction on Waymo validation set.}
    \label{fig:occ_vis}
\end{figure}

\subsubsection{Ablation Studies of \method} 

\noindent \textbf{Module-level Studies.} We conduct ablation experiments to analyze the individual components of the proposed \method. For pre-training, we uniformly sample $5\%$ Waymo data and subsequently perform fine-tuning experiments on subsets of $5\%$ nuScenes (det) data, $20\%$ KITTI data, and $20\%$ ONCE dataset, using SECOND~\cite{second} as the detector. The results presented in Table~\ref{tab:abl_results} demonstrate the effectiveness of the proposed occupancy prediction task in enhancing the performance of the downstream tasks. Moreover, our proposed strategies for pre-training, including loss balancing, beam re-sampling, and dataset balancing, yield significant improvements in different datasets.

\noindent \textbf{Generalizability Studies.}
To further verify the generalizability of our approach towards the Transformer-based network structure, we have conducted experiments on DSVT model~\cite{wang2023dsvt}. First, we employ the encoder of DSVT model~\cite{wang2023dsvt} and perform the pre-training process using SPOT on 20\% sequence-level data from Waymo. Then, the fine-tuning experiments are conducted on the nuScenes and ONCE datasets. The results shown in Table~\ref{tab:DSVT_results} demonstrate that, for the transformer-based baseline, SPOT also achieves significant gains under different benchmarks.


\subsubsection{Discussions of Downstream Tasks}

\noindent \textbf{Data-Efficiency for Downstream.} In order to illustrate the influence of the pre-training method on downstream data, we conduct the fine-tuning experiments on nuScenes dataset using varying proportions of annotated data (\textit{e.g.}, 5\%, 10\%, 25\%, and 100\% budgets), using SECOND~\cite{second} as the detector. Fig.~\ref{fig:nusc_scale_results} shows the results of our experiments, highlighting the consistent performance improvement achieved by \method\ across different budget allocations, demonstrating its effectiveness in improving data efficiency.

\noindent \textbf{Fine-tuning Performance on Waymo Detection.} We perform detailed experiments in the downstream Waymo detection task. We evaluate the results using the official Average Precision (AP) and Average Precision with Heading (APH), with a  focus on the more challenging L2-LEVEL metrics. The evaluation results on the Waymo validation set are presented in Table~\ref{tab:waymo_results}. We conduct fine-tuning on $3\%$ data using the widely adopted CenterPoint detector~\cite{centerpoint}. Furthermore, we confirm the scalability of \method\ and achieve superior performance compared to training from scratch. Specifically, \method\ improves the performance of training from scratch by 4.76\% and 4.69\% for CenterPoint in L2 AP and L2 APH. Table~\ref{tab:waymo_results} illustrates that \method\ with only 5\% sequence-level pre-training data can outperform BEV-MAE~\cite{bev-mae} and AD-PT~\cite{ad-pt} using 100\% pre-training data.

\noindent \textbf{Beyond the Label-efficiency Downstream Setting.} We further conduct experiments on complete downstream datasets, \textit{i.e.}, using 100\% training data from downstream tasks to conduct the fine-tuning. The results are shown in Table~\ref{tab:det_100_results} and Table~\ref{tab:seg_100_results}. It can be found that \method\ also achieves consistent performance gains even with $100\%$ labeled data for fine-tuning, which highlights the effectiveness of \method.



\subsection{Visualization Results}
\label{append:vis}

Firstly, Fig.~\ref{fig:det_vis} shows the visualization results of different downstream datasets (\textit{i.e.}, KITTI, ONCE). The visualization results of different downstream datasets also demonstrate that our SPOT boosts the ability of the baseline for 3D object detection task compared to training from scratch.

Secondly, Fig.~\ref{fig:occ_vis} visualizes the results obtained from our pre-training task on the Waymo validation set, showcasing the raw input point cloud on the left, while the middle and right sections display our predicted occupancy results and the Ground Truth (GT) of the dataset, respectively. Fig.~\ref{fig:occ_vis} clearly demonstrates our ability to generate highly dense occupancy prediction using a sparse single-frame point cloud input. Furthermore, it is worth noting that the occupancy GT also exhibits sparsity in certain areas, such as certain sections of the road surface. This sparsity is inherent to LiDAR sensor, as there will always be some areas that are not scanned and virtually have no points in the frame. However, our prediction results exhibit greater continuity and produce superior performance in these details, which confirms the scene understanding capability of \method.

{\color{red}
\subsection{Limitations and Future Directions}
\label{append:limitation}

While SPOT demonstrates promising results across multiple datasets and tasks, several limitations warrant discussion:

\textbf{Sensor Alignment Challenges:} The practical deployment of SPOT faces engineering challenges related to camera-LiDAR calibration. While modern autonomous driving systems typically require such calibration for basic operation, achieving the precision necessary for high-quality pseudo-label generation remains a significant engineering investment, particularly for organizations without established multi-modal data pipelines.

\textbf{Single-Dataset Pre-training Scope:} Our current framework focuses on single-dataset pre-training to establish foundational transferability. While Fig.~\ref{fig:once_results} suggests that more pre-training data could yield further improvements, multi-dataset joint pre-training introduces additional complexities including dataset fusion strategies, domain mixing effects, and potential negative transfer that require systematic investigation.
}

\section{Conclusion}
In this paper, we have introduced \method, a scalable and general 3D representation learning method for LiDAR point clouds. \method\ utilizes occupancy prediction as the pre-training task and narrows domain gaps between different datasets by beam re-sampling augmentation and class-balancing strategies. Besides, we conduct a thorough theoretical analysis to uncover why the proposed occupancy pre-training task obtains temporally sufficient representations. Experimentally, consistent improvement in various downstream datasets and tasks as well as scalable pre-training are observed. We believe \method\ paves the way for large-scale pre-training on LiDAR point clouds.

\section*{Acknowledgements}
The research was supported by Shanghai Artificial Intelligence Laboratory, the National Key R\&D Program of China (Grant No. 2022ZD0160104), and the Science and Technology Commission of Shanghai Municipality (Grant No. 22DZ1100102). This work was supported by a locally commissioned task from the Shanghai Municipal Government, Shanghai Rising Star Program (Grant No. 23QD1401000), and NSFC (72342023).

\ifCLASSOPTIONcaptionsoff
  \newpage
\fi



%
\bibliographystyle{IEEEtran}
\bibliography{pami}

\begin{thebibliography}{10}
\providecommand{\url}[1]{#1}
\csname url@samestyle\endcsname
\providecommand{\newblock}{\relax}
\providecommand{\bibinfo}[2]{#2}
\providecommand{\BIBentrySTDinterwordspacing}{\spaceskip=0pt\relax}
\providecommand{\BIBentryALTinterwordstretchfactor}{4}
\providecommand{\BIBentryALTinterwordspacing}{\spaceskip=\fontdimen2\font plus
\BIBentryALTinterwordstretchfactor\fontdimen3\font minus \fontdimen4\font\relax}
\providecommand{\BIBforeignlanguage}[2]{{%
\expandafter\ifx\csname l@#1\endcsname\relax
\typeout{** WARNING: IEEEtran.bst: No hyphenation pattern has been}%
\typeout{** loaded for the language `#1'. Using the pattern for}%
\typeout{** the default language instead.}%
\else
\language=\csname l@#1\endcsname
\fi
#2}}
\providecommand{\BIBdecl}{\relax}
\BIBdecl

\bibitem{pmlr-v164-jia22a}
X.~Jia, L.~Sun, H.~Zhao, M.~Tomizuka, and W.~Zhan, ``Multi-agent trajectory prediction by combining egocentric and allocentric views,'' in \emph{5th Conference on Robot Learning}, ser. Proceedings of Machine Learning Research, vol. 164.\hskip 1em plus 0.5em minus 0.4em\relax PMLR, 08--11 Nov 2022, pp. 1434--1443.

\bibitem{pmlr-v205-jia23a}
X.~Jia, L.~Chen, P.~Wu, J.~Zeng, J.~Yan, H.~Li, and Y.~Qiao, ``Towards capturing the temporal dynamics for trajectory prediction: a coarse-to-fine approach,'' in \emph{6th Conference on Robot Learning}, ser. Proceedings of Machine Learning Research, vol. 205, 2023, pp. 910--920.

\bibitem{raljia}
X.~Jia, L.~Sun, M.~Tomizuka, and W.~Zhan, ``Ide-net: Interactive driving event and pattern extraction from human data,'' \emph{IEEE Robotics and Automation Letters}, vol.~6, no.~2, pp. 3065--3072, 2021.

\bibitem{Jia2022HDGTHD}
X.~Jia, P.~Wu, L.~Chen, H.~Li, Y.~S. Liu, and J.~Yan, ``Hdgt: Heterogeneous driving graph transformer for multi-agent trajectory prediction via scene encoding,'' \emph{IEEE Transactions on Pattern Analysis and Machine Intelligence}, vol.~45, pp. 13\,860--13\,875, 2022.

\bibitem{jia2024amp}
X.~Jia, S.~Shi, Z.~Chen, L.~Jiang, W.~Liao, T.~He, and J.~Yan, ``Amp: Autoregressive motion prediction revisited with next token prediction for autonomous driving,'' 2024.

\bibitem{wu2022trajectory}
P.~Wu, X.~Jia, L.~Chen, J.~Yan, H.~Li, and Y.~Qiao, ``Trajectory-guided control prediction for end-to-end autonomous driving: A simple yet strong baseline,'' in \emph{Advances in Neural Information Processing Systems}, vol.~35, 2022, pp. 6119--6132.

\bibitem{jia2023think}
X.~Jia, P.~Wu, L.~Chen, J.~Xie, C.~He, J.~Yan, and H.~Li, ``Think twice before driving: Towards scalable decoders for end-to-end autonomous driving,'' in \emph{IEEE/CVF Conference on Computer Vision and Pattern Recognition}, 2023, pp. 21\,983--21\,994.

\bibitem{jia2023driveadapter}
X.~Jia, Y.~Gao, L.~Chen, J.~Yan, P.~L. Liu, and H.~Li, ``Driveadapter: Breaking the coupling barrier of perception and planning in end-to-end autonomous driving,'' in \emph{IEEE/CVF International Conference on Computer Vision}, 2023, pp. 7953--7963.

\bibitem{yang2023survey}
Z.~Yang, X.~Jia, H.~Li, and J.~Yan, ``Llm4drive: A survey of large language models for autonomous driving,'' \emph{arXiv preprint arXiv:2311.01043}, 2023.

\bibitem{li2024think}
Q.~Li, X.~Jia, S.~Wang, and J.~Yan, ``Think2drive: Efficient reinforcement learning by thinking in latent world model for quasi-realistic autonomous driving (in carla-v2),'' in \emph{ECCV}, 2024.

\bibitem{second}
Y.~Yan, Y.~Mao, and B.~Li, ``Second: Sparsely embedded convolutional detection,'' \emph{Sensors}, vol.~18, no.~10, p. 3337, 2018.

\bibitem{centerpoint}
T.~Yin, X.~Zhou, and P.~Krahenbuhl, ``Center-based 3d object detection and tracking,'' in \emph{IEEE/CVF Conference on Computer Vision and Pattern Recognition}, 2021, pp. 11\,784--11\,793.

\bibitem{pv-rcnn}
S.~Shi, C.~Guo, L.~Jiang, Z.~Wang, J.~Shi, X.~Wang, and H.~Li, ``Pv-rcnn: Point-voxel feature set abstraction for 3d object detection,'' in \emph{IEEE/CVF Conference on Computer Vision and Pattern Recognition}, 2020, pp. 10\,529--10\,538.

\bibitem{pv-rcnn++}
S.~Shi, L.~Jiang, J.~Deng, Z.~Wang, C.~Guo, J.~Shi, X.~Wang, and H.~Li, ``Pv-rcnn++: Point-voxel feature set abstraction with local vector representation for 3d object detection,'' \emph{International Journal of Computer Vision}, vol. 131, no.~2, pp. 531--551, 2023.

\bibitem{cylinder3d}
X.~Zhu, H.~Zhou, T.~Wang, F.~Hong, Y.~Ma, W.~Li, H.~Li, and D.~Lin, ``Cylindrical and asymmetrical 3d convolution networks for lidar segmentation,'' in \emph{IEEE Conference on Computer Vision and Pattern Recognition}, 2021, pp. 9939--9948.

\bibitem{uni3d}
B.~Zhang, J.~Yuan, B.~Shi, T.~Chen, Y.~Li, and Y.~Qiao, ``Uni3d: A unified baseline for multi-dataset 3d object detection,'' in \emph{IEEE/CVF Conference on Computer Vision and Pattern Recognition}, 2023, pp. 9253--9262.

\bibitem{yuan2023bi3d}
J.~Yuan, B.~Zhang, X.~Yan, T.~Chen, B.~Shi, Y.~Li, and Y.~Qiao, ``Bi3d: Bi-domain active learning for cross-domain 3d object detection,'' in \emph{IEEE/CVF Conference on Computer Vision and Pattern Recognition}, 2023, pp. 15\,599--15\,608.

\bibitem{kitti}
A.~Geiger, P.~Lenz, and R.~Urtasun, ``Are we ready for autonomous driving? the kitti vision benchmark suite,'' in \emph{IEEE/CVF Conference on Computer Vision and Pattern Recognition}, 2012, pp. 3354--3361.

\bibitem{semantickitti}
J.~Behley, M.~Garbade, A.~Milioto, J.~Quenzel, S.~Behnke, C.~Stachniss, and J.~Gall, ``Semantickitti: A dataset for semantic scene understanding of lidar sequences,'' in \emph{IEEE/CVF international conference on computer vision}, 2019, pp. 9297--9307.

\bibitem{once}
J.~Mao, M.~Niu, C.~Jiang, H.~Liang, J.~Chen, X.~Liang, Y.~Li, C.~Ye, W.~Zhang, Z.~Li \emph{et~al.}, ``One million scenes for autonomous driving: Once dataset,'' \emph{arXiv preprint arXiv:2106.11037}, 2021.

\bibitem{nuscenes}
H.~Caesar, V.~Bankiti, A.~H. Lang, S.~Vora, V.~E. Liong, Q.~Xu, A.~Krishnan, Y.~Pan, G.~Baldan, and O.~Beijbom, ``nuscenes: A multimodal dataset for autonomous driving,'' in \emph{IEEE/CVF Conference on Computer Vision and Pattern Recognition}, 2020, pp. 11\,621--11\,631.

\bibitem{waymo}
P.~Sun, H.~Kretzschmar, X.~Dotiwalla, A.~Chouard, V.~Patnaik, P.~Tsui, J.~Guo, Y.~Zhou, Y.~Chai, B.~Caine \emph{et~al.}, ``Scalability in perception for autonomous driving: Waymo open dataset,'' in \emph{IEEE/CVF Conference on Computer Vision and Pattern Recognition}, 2020, pp. 2446--2454.

\bibitem{jia2024bench2drive}
X.~Jia, Z.~Yang, Q.~Li, Z.~Zhang, and J.~Yan, ``Bench2drive: Towards multi-ability benchmarking of closed-loop end-to-end autonomous driving,'' in \emph{NeurIPS}, 2024.

\bibitem{lu2024activead}
H.~Lu, X.~Jia, Y.~Xie, W.~Liao, X.~Yang, and J.~Yan, ``Activead: Planning-oriented active learning for end-to-end autonomous driving,'' \emph{arXiv preprint arXiv:2403.02877}, 2024.

\bibitem{ad-pt}
J.~Yuan, B.~Zhang, X.~Yan, B.~Shi, T.~Chen, Y.~Li, and Y.~Qiao, ``Ad-pt: Autonomous driving pre-training with large-scale point cloud dataset,'' \emph{Advances in Neural Information Processing Systems}, vol.~36, 2024.

\bibitem{gcc-3d}
H.~Liang, C.~Jiang, D.~Feng, X.~Chen, H.~Xu, X.~Liang, W.~Zhang, Z.~Li, and L.~Van~Gool, ``Exploring geometry-aware contrast and clustering harmonization for self-supervised 3d object detection,'' in \emph{IEEE/CVF International Conference on Computer Vision}, 2021, pp. 3293--3302.

\bibitem{strl}
S.~Huang, Y.~Xie, S.-C. Zhu, and Y.~Zhu, ``Spatio-temporal self-supervised representation learning for 3d point clouds,'' in \emph{IEEE/CVF International Conference on Computer Vision}, 2021, pp. 6535--6545.

\bibitem{bev-mae}
Z.~Lin, Y.~Wang, S.~Qi, N.~Dong, and M.-H. Yang, ``Bev-mae: Bird’s eye view masked autoencoders for point cloud pre-training in autonomous driving scenarios,'' in \emph{AAAI Conference on Artificial Intelligence}, vol.~38, no.~4, 2024, pp. 3531--3539.

\bibitem{co3}
R.~Chen, Y.~Mu, R.~Xu, W.~Shao, C.~Jiang, H.~Xu, Z.~Li, and P.~Luo, ``Co\^{} 3: Cooperative unsupervised 3d representation learning for autonomous driving,'' \emph{arXiv preprint arXiv:2206.04028}, 2022.

\bibitem{gdmae}
H.~Yang, T.~He, J.~Liu, H.~Chen, B.~Wu, B.~Lin, X.~He, and W.~Ouyang, ``Gd-mae: generative decoder for mae pre-training on lidar point clouds,'' in \emph{IEEE/CVF Conference on Computer Vision and Pattern Recognition}, 2023, pp. 9403--9414.

\bibitem{pointrcnn}
S.~Shi, X.~Wang, and H.~Li, ``Pointrcnn: 3d object proposal generation and detection from point cloud,'' in \emph{IEEE/CVF Conference on Computer Vision and Pattern Recognition (CVPR)}, June 2019.

\bibitem{3dssd}
Z.~Yang, Y.~Sun, S.~Liu, and J.~Jia, ``3dssd: Point-based 3d single stage object detector,'' in \emph{IEEE/CVF conference on computer vision and pattern recognition}, 2020, pp. 11\,040--11\,048.

\bibitem{pointformer}
X.~Pan, Z.~Xia, S.~Song, L.~E. Li, and G.~Huang, ``3d object detection with pointformer,'' in \emph{IEEE/CVF conference on computer vision and pattern recognition}, 2021, pp. 7463--7472.

\bibitem{voxelnet}
Y.~Zhou and O.~Tuzel, ``Voxelnet: End-to-end learning for point cloud based 3d object detection,'' in \emph{IEEE Conference on Computer Vision and Pattern Recognition}, 2018.

\bibitem{part-a2}
S.~Shi, Z.~Wang, J.~Shi, X.~Wang, and H.~Li, ``From points to parts: 3d object detection from point cloud with part-aware and part-aggregation network,'' \emph{IEEE transactions on pattern analysis and machine intelligence}, vol.~43, no.~8, pp. 2647--2664, 2020.

\bibitem{voxel-rcnn}
J.~Deng, S.~Shi, P.~Li, W.~Zhou, Y.~Zhang, and H.~Li, ``Voxel r-cnn: Towards high performance voxel-based 3d object detection,'' in \emph{AAAI conference on artificial intelligence}, vol.~35, no.~2, 2021, pp. 1201--1209.

\bibitem{fast_pointrcnn}
Y.~Chen, S.~Liu, X.~Shen, and J.~Jia, ``Fast point r-cnn,'' in \emph{IEEE/CVF international conference on computer vision}, 2019, pp. 9775--9784.

\bibitem{lidar-rcnn}
Z.~Li, F.~Wang, and N.~Wang, ``Lidar r-cnn: An efficient and universal 3d object detector,'' in \emph{IEEE/CVF conference on computer vision and pattern recognition}, 2021, pp. 7546--7555.

\bibitem{scribble}
O.~Unal, D.~Dai, and L.~Van~Gool, ``Scribble-supervised lidar semantic segmentation,'' in \emph{IEEE conference on computer vision and pattern recognition}, 2022, pp. 2697--2707.

\bibitem{lasermix}
L.~Kong, J.~Ren, L.~Pan, and Z.~Liu, ``Lasermix for semi-supervised lidar semantic segmentation,'' in \emph{IEEE conference on computer vision and pattern recognition}, 2023, pp. 21\,705--21\,715.

\bibitem{Lim3D}
L.~Li, H.~P. Shum, and T.~P. Breckon, ``Less is more: Reducing task and model complexity for 3d point cloud semantic segmentation,'' in \emph{IEEE conference on computer vision and pattern recognition}, 2023, pp. 9361--9371.

\bibitem{xie2020pointcontrast}
S.~Xie, J.~Gu, D.~Guo, C.~R. Qi, L.~Guibas, and O.~Litany, ``Pointcontrast: Unsupervised pre-training for 3d point cloud understanding,'' in \emph{Computer Vision--ECCV 2020: 16th European Conference, Glasgow, UK, August 23--28, 2020, Proceedings, Part III 16}.\hskip 1em plus 0.5em minus 0.4em\relax Springer, 2020, pp. 574--591.

\bibitem{yin2022proposalcontrast}
J.~Yin, D.~Zhou, L.~Zhang, J.~Fang, C.-Z. Xu, J.~Shen, and W.~Wang, ``Proposalcontrast: Unsupervised pre-training for lidar-based 3d object detection,'' in \emph{European conference on computer vision}.\hskip 1em plus 0.5em minus 0.4em\relax Springer, 2022, pp. 17--33.

\bibitem{min2024driveworld}
C.~Min, D.~Zhao, L.~Xiao, J.~Zhao, X.~Xu, Z.~Zhu, L.~Jin, J.~Li, Y.~Guo, J.~Xing \emph{et~al.}, ``Driveworld: 4d pre-trained scene understanding via world models for autonomous driving,'' in \emph{IEEE/CVF Conference on Computer Vision and Pattern Recognition}, 2024, pp. 15\,522--15\,533.

\bibitem{occ-mae}
C.~Min, L.~Xiao, D.~Zhao, Y.~Nie, and B.~Dai, ``Occupancy-mae: Self-supervised pre-training large-scale lidar point clouds with masked occupancy autoencoders,'' \emph{IEEE Transactions on Intelligent Vehicles}, 2023.

\bibitem{mv-jar}
R.~Xu, T.~Wang, W.~Zhang, R.~Chen, J.~Cao, J.~Pang, and D.~Lin, ``Mv-jar: Masked voxel jigsaw and reconstruction for lidar-based self-supervised pre-training,'' in \emph{IEEE/CVF Conference on Computer Vision and Pattern Recognition}, 2023, pp. 13\,445--13\,454.

\bibitem{monoscene}
A.-Q. Cao and R.~de~Charette, ``Monoscene: Monocular 3d semantic scene completion,'' in \emph{IEEE/CVF Conference on Computer Vision and Pattern Recognition}, 2022, pp. 3991--4001.

\bibitem{voxformer}
Y.~Li, Z.~Yu, C.~Choy, C.~Xiao, J.~M. Alvarez, S.~Fidler, C.~Feng, and A.~Anandkumar, ``Voxformer: Sparse voxel transformer for camera-based 3d semantic scene completion,'' in \emph{IEEE/CVF Conference on Computer Vision and Pattern Recognition}, 2023, pp. 9087--9098.

\bibitem{tpvformer}
Y.~Huang, W.~Zheng, Y.~Zhang, J.~Zhou, and J.~Lu, ``Tri-perspective view for vision-based 3d semantic occupancy prediction,'' in \emph{IEEE/CVF Conference on Computer Vision and Pattern Recognition}, 2023, pp. 9223--9232.

\bibitem{js3c-net}
X.~Yan, J.~Gao, J.~Li, R.~Zhang, Z.~Li, R.~Huang, and S.~Cui, ``Sparse single sweep lidar point cloud segmentation via learning contextual shape priors from scene completion,'' in \emph{AAAI Conference on Artificial Intelligence}, vol.~35, no.~4, 2021, pp. 3101--3109.

\bibitem{scpnet}
Z.~Xia, Y.~Liu, X.~Li, X.~Zhu, Y.~Ma, Y.~Li, Y.~Hou, and Y.~Qiao, ``Scpnet: Semantic scene completion on point cloud,'' in \emph{IEEE/CVF Conference on Computer Vision and Pattern Recognition}, 2023, pp. 17\,642--17\,651.

\bibitem{wang2023openoccupancy}
X.~Wang, Z.~Zhu, W.~Xu, Y.~Zhang, Y.~Wei, X.~Chi, Y.~Ye, D.~Du, J.~Lu, and X.~Wang, ``Openoccupancy: A large scale benchmark for surrounding semantic occupancy perception,'' in \emph{IEEE/CVF International Conference on Computer Vision}, 2023, pp. 17\,850--17\,859.

\bibitem{zhang2023occformer}
Y.~Zhang, Z.~Zhu, and D.~Du, ``Occformer: Dual-path transformer for vision-based 3d semantic occupancy prediction,'' in \emph{IEEE/CVF International Conference on Computer Vision}, 2023, pp. 9433--9443.

\bibitem{ma2024cotr}
Q.~Ma, X.~Tan, Y.~Qu, L.~Ma, Z.~Zhang, and Y.~Xie, ``Cotr: Compact occupancy transformer for vision-based 3d occupancy prediction,'' in \emph{IEEE/CVF Conference on Computer Vision and Pattern Recognition}, 2024, pp. 19\,936--19\,945.

\bibitem{pan2023uniocc}
M.~Pan, L.~Liu, J.~Liu, P.~Huang, L.~Wang, S.~Zhang, S.~Xu, Z.~Lai, and K.~Yang, ``Uniocc: Unifying vision-centric 3d occupancy prediction with geometric and semantic rendering,'' \emph{arXiv preprint arXiv:2306.09117}, 2023.

\bibitem{vobecky2024pop}
A.~Vobecky, O.~Sim{\'e}oni, D.~Hurych, S.~Gidaris, A.~Bursuc, P.~P{\'e}rez, and J.~Sivic, ``Pop-3d: Open-vocabulary 3d occupancy prediction from images,'' \emph{Advances in Neural Information Processing Systems}, vol.~36, 2024.

\bibitem{tang2024sparseocc}
P.~Tang, Z.~Wang, G.~Wang, J.~Zheng, X.~Ren, B.~Feng, and C.~Ma, ``Sparseocc: Rethinking sparse latent representation for vision-based semantic occupancy prediction,'' in \emph{IEEE/CVF Conference on Computer Vision and Pattern Recognition}, 2024, pp. 15\,035--15\,044.

\bibitem{li2024pmafusion}
S.~Li, W.~Yang, and Q.~Liao, ``Pmafusion: Projection-based multi-modal alignment for 3d semantic occupancy prediction,'' in \emph{IEEE/CVF Conference on Computer Vision and Pattern Recognition}, 2024, pp. 3627--3634.

\bibitem{zhao2024lowrankocc}
L.~Zhao, X.~Xu, Z.~Wang, Y.~Zhang, B.~Zhang, W.~Zheng, D.~Du, J.~Zhou, and J.~Lu, ``Lowrankocc: Tensor decomposition and low-rank recovery for vision-based 3d semantic occupancy prediction,'' in \emph{IEEE/CVF Conference on Computer Vision and Pattern Recognition}, 2024, pp. 9806--9815.

\bibitem{huang2024selfocc}
Y.~Huang, W.~Zheng, B.~Zhang, J.~Zhou, and J.~Lu, ``Selfocc: Self-supervised vision-based 3d occupancy prediction,'' in \emph{IEEE/CVF Conference on Computer Vision and Pattern Recognition}, 2024, pp. 19\,946--19\,956.

\bibitem{occ3d}
X.~Tian, T.~Jiang, L.~Yun, Y.~Mao, H.~Yang, Y.~Wang, Y.~Wang, and H.~Zhao, ``Occ3d: A large-scale 3d occupancy prediction benchmark for autonomous driving,'' \emph{Advances in Neural Information Processing Systems}, vol.~36, 2024.

\bibitem{lovasz}
M.~Berman, A.~R. Triki, and M.~B. Blaschko, ``The lov{\'a}sz-softmax loss: A tractable surrogate for the optimization of the intersection-over-union measure in neural networks,'' in \emph{CVPR}, 2018, pp. 4413--4421.

\bibitem{class-balancing-sampling}
B.~Zhu, Z.~Jiang, X.~Zhou, Z.~Li, and G.~Yu, ``Class-balanced grouping and sampling for point cloud 3d object detection,'' \emph{arXiv preprint arXiv:1908.09492}, 2019.

\bibitem{rethinking}
H.~Wang, X.~Guo, Z.-H. Deng, and Y.~Lu, ``Rethinking minimal sufficient representation in contrastive learning,'' in \emph{IEEE/CVF Conference on Computer Vision and Pattern Recognition}, 2022, pp. 16\,041--16\,050.

\bibitem{statistical}
K.~Fukunaga, \emph{Introduction to statistical pattern recognition}.\hskip 1em plus 0.5em minus 0.4em\relax Elsevier, 2013.

\bibitem{zhang2023resimad}
B.~Zhang, X.~Cai, J.~Yuan, D.~Yang, J.~Guo, R.~Xia, B.~Shi, M.~Dou, T.~Chen, S.~Liu \emph{et~al.}, ``Resimad: Zero-shot 3d domain transfer for autonomous driving with source reconstruction and target simulation,'' \emph{arXiv preprint arXiv:2309.05527}, 2023.

\bibitem{3DTrans}
D.~D. Team, ``3dtrans: An open-source codebase for exploring transferable autonomous driving perception task,'' \url{https://github.com/PJLab-ADG/3DTrans}, 2023.

\bibitem{pseudo-label}
D.-H. Lee \emph{et~al.}, ``Pseudo-label: The simple and efficient semi-supervised learning method for deep neural networks,'' in \emph{Workshop on challenges in representation learning, ICML}, vol.~3, no.~2, 2013, p. 896.

\bibitem{mean-teacher}
A.~Tarvainen and H.~Valpola, ``Mean teachers are better role models: Weight-averaged consistency targets improve semi-supervised deep learning results,'' in \emph{Advances in neural information processing systems}, vol.~30, 2017.

\bibitem{ravi2024sam}
N.~Ravi, V.~Gabeur, Y.-T. Hu, R.~Hu, C.~Ryali, T.~Ma, H.~Khedr, R.~R{\"a}dle, C.~Rolland, L.~Gustafson \emph{et~al.}, ``Sam 2: Segment anything in images and videos,'' \emph{arXiv preprint arXiv:2408.00714}, 2024.

\bibitem{liu2025grounding}
S.~Liu, Z.~Zeng, T.~Ren, F.~Li, H.~Zhang, J.~Yang, Q.~Jiang, C.~Li, J.~Yang, H.~Su \emph{et~al.}, ``Grounding dino: Marrying dino with grounded pre-training for open-set object detection,'' in \emph{ECCV}, 2025, pp. 38--55.

\bibitem{wang2023dsvt}
H.~Wang, C.~Shi, S.~Shi, M.~Lei, S.~Wang, D.~He, B.~Schiele, and L.~Wang, ``Dsvt: Dynamic sparse voxel transformer with rotated sets,'' in \emph{IEEE/CVF Conference on Computer Vision and Pattern Recognition}, 2023, pp. 13\,520--13\,529.

\end{thebibliography}

\newpage


\begin{IEEEbiography}[{\includegraphics[width=1in,height=1.25in,clip,keepaspectratio]{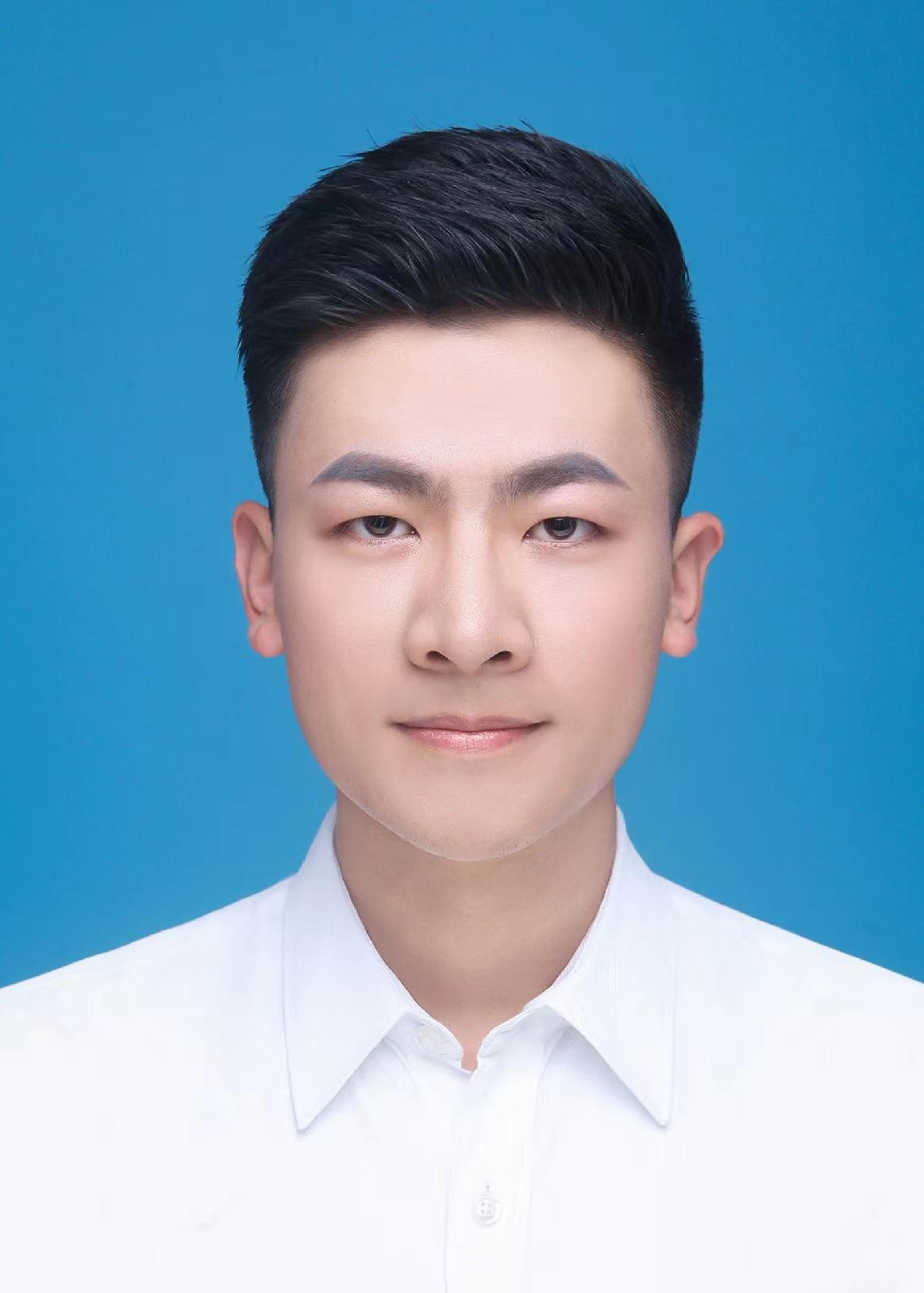}}]{Xiangchao Yan}
recieved M.S. degree from Shanghai Jiao Tong University. He is currently a Researcher at Shanghai Artificial Intelligence Laboratory. His research interests include generalized representation learning in 3D scenes, 3D scene pre-training, and multi-modal document understanding.
\end{IEEEbiography}


\begin{IEEEbiography}[{\includegraphics[width=1in,height=1.25in,clip,keepaspectratio]{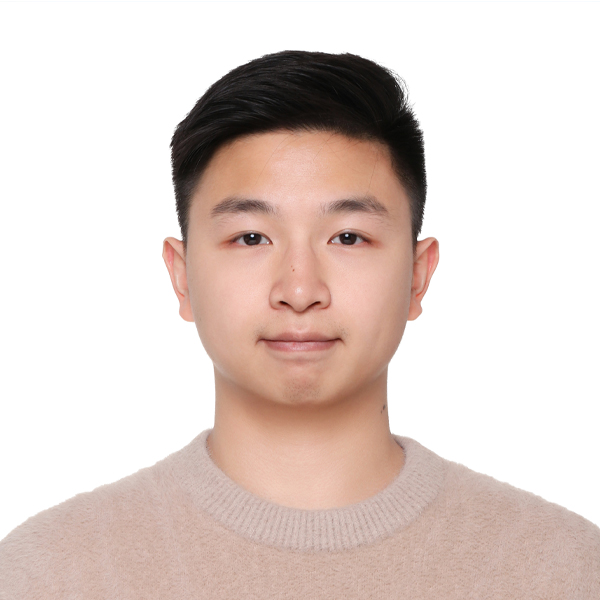}}]{Runjian Chen}
received his B.S. degree in Automation (Robotics track) from Zhejiang University with an honor degree for the Mixed Class at Chu Kochen Honor College. He is currently a Ph.D student at HKU-MMLab. His research focus is on unsupervised/semi-supervised 3D representation learning and its application in robotics and autonomous driving.
\end{IEEEbiography}



\begin{IEEEbiography}[{\includegraphics[width=1in,height=1.25in,clip,keepaspectratio]{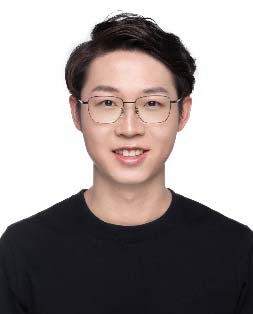}}]{Bo Zhang} is a Research Scientist at Shanghai Artificial Intelligence Laboratory. His work has garnered awards, e.g. Shanghai Rising Star. He led the development of the 3DTrans general scene representation open-source project, which won the Waymo Challenge  and accumulated over 10k stars. He also works on the application of multi-modal large language models in various scenarios, e.g. AI for scientific discovery and reasoning.
\end{IEEEbiography}

\begin{IEEEbiography}[{\includegraphics[width=1in,height=1.25in,clip,keepaspectratio]{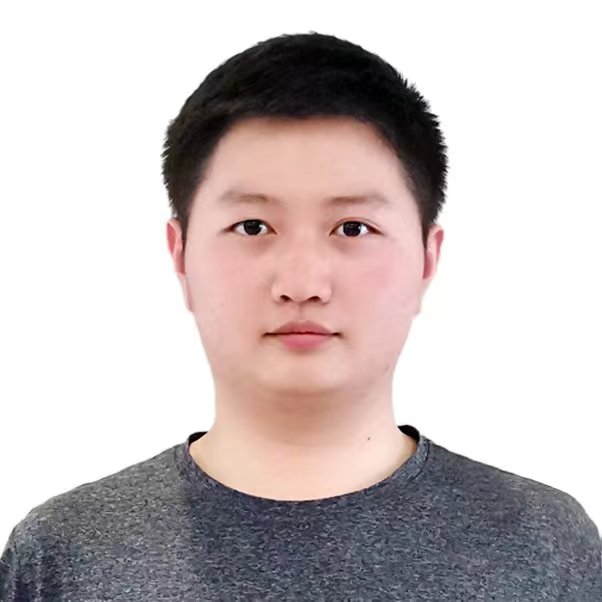}}]{Hancheng Ye}
received his B.S. and M.S. degrees in Electronic Engineering at School of Information Science and Technology from Fudan University. He is currently a Research Assistant at Shanghai Artificial Intelligence Laboratory. His primary research interests focus on efficient machine learning, model compression, and multimodal learning.
\end{IEEEbiography}


\begin{IEEEbiography}[{\includegraphics[width=1in,height=1.25in,clip,keepaspectratio]{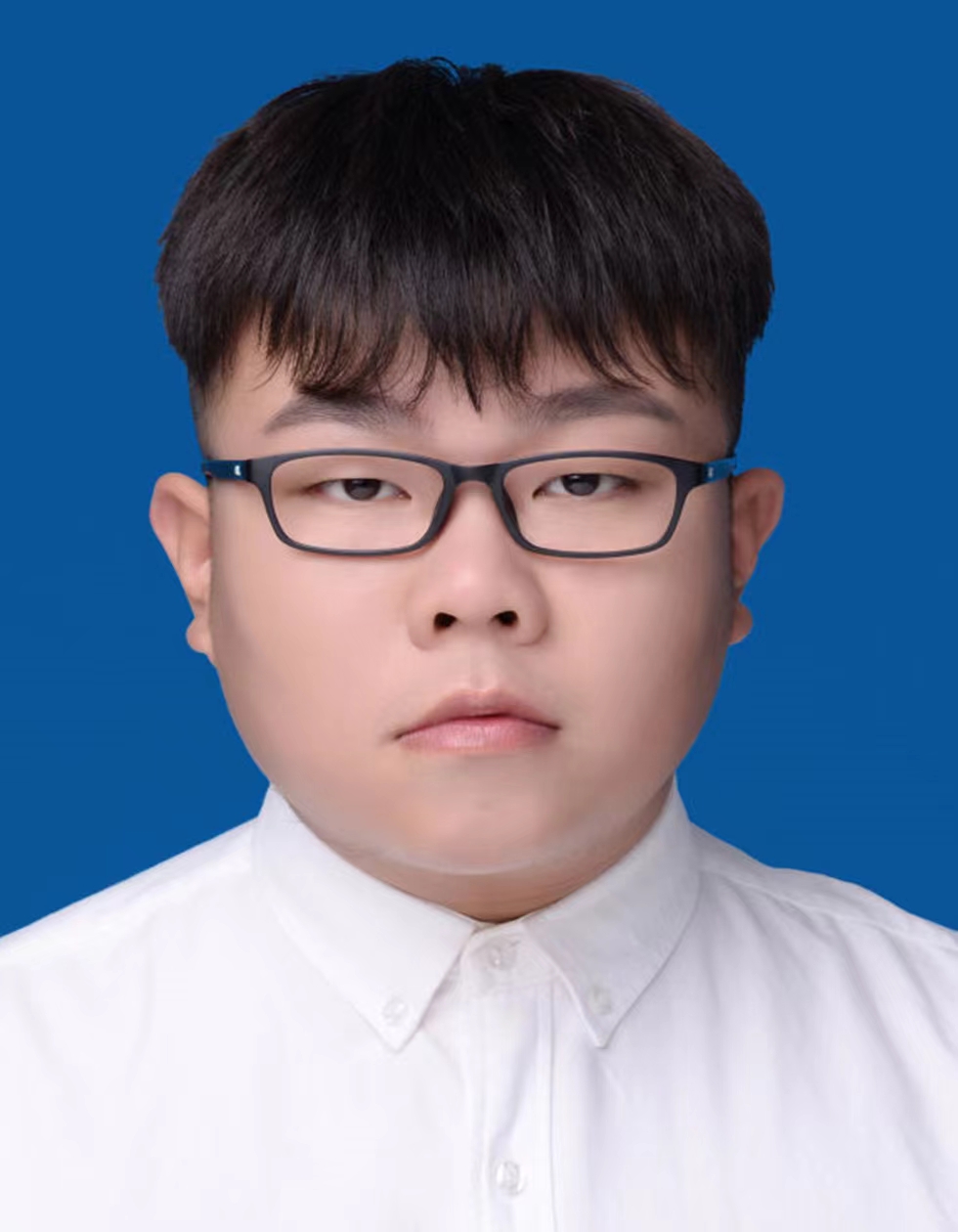}}]{Renqiu Xia}
is an Assistant Professor with School of Artificial Intelligence, Shanghai Jiao Tong University. His primary research interests encompass Neural Architecture Search and Multi-modal Large Language Models, with a particular focus on chart and document understanding as well as mathematical reasoning. He regularly serves as a reviewer for KDD and TKDE.
\end{IEEEbiography}


\begin{IEEEbiography}[{\includegraphics[width=1in,height=1.25in,clip,keepaspectratio]{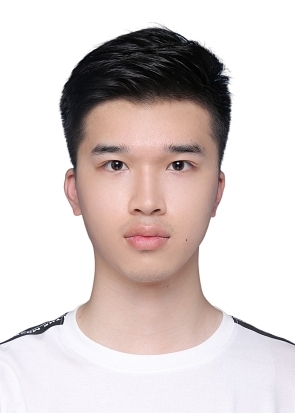}}]{Jiakang Yuan}
is currently a Ph.D. student in Electronic Engineering at School of Information Science and Technology, Fudan University (2022 - 2027).
Before that, he received his bachelor's degree in Electronic Engineering also from Fudan University (2018 - 2022).
His research interests include multimodal reasoning, multi-agent system, and spatial intelligence.
\end{IEEEbiography}


\begin{IEEEbiography}[{\includegraphics[width=1in,height=1.25in,clip,keepaspectratio]{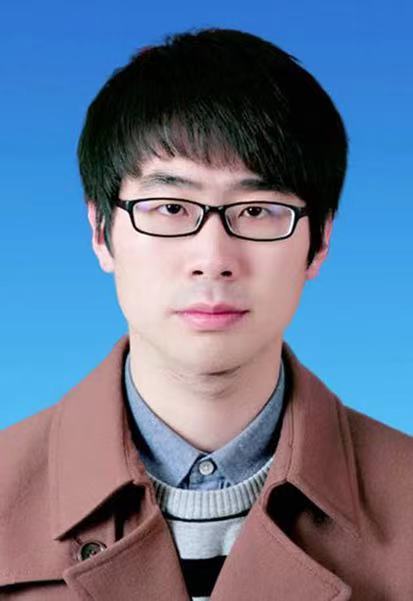}}]{Hongbin Zhou} received M.S. degree from Xi'an Jiaotong University, 
and is currently a Researcher at Shanghai Artificial Intelligence Laboratory. His research interests
include autonomous driving and computer vision.
\end{IEEEbiography}


\begin{IEEEbiography}[{\includegraphics[width=1in,height=1.25in,clip,keepaspectratio]{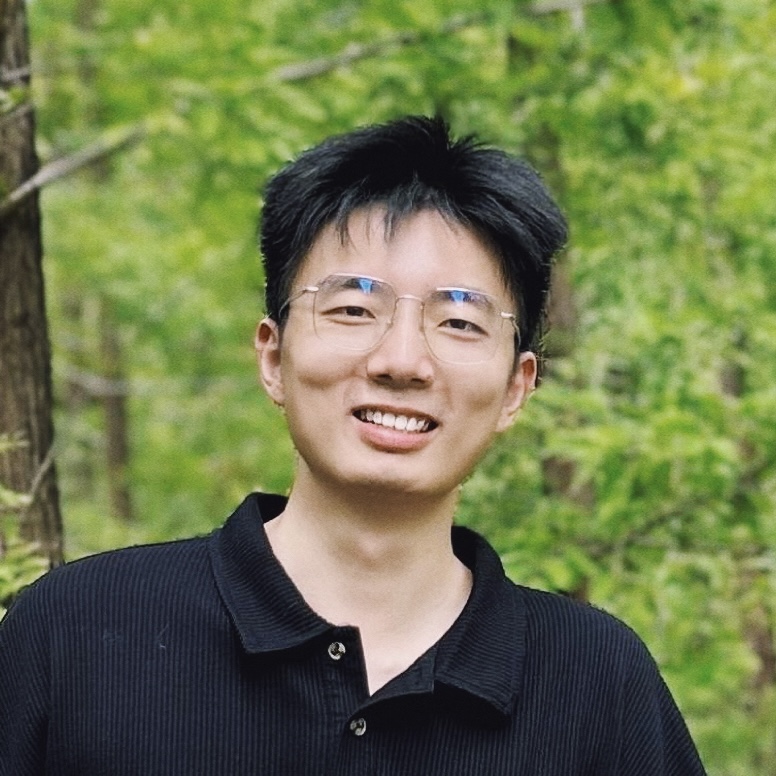}}]{Xinyu Cai} is a Researcher currently affiliated at Shanghai Artificial Intelligence Laboratory. He received M.S. degree from the University of Chinese Academy of Sciences. His research interests focus on multi-modal large models and real-scene generation.
\end{IEEEbiography}


\begin{IEEEbiography}[{\includegraphics[width=1in,height=1.25in,clip,keepaspectratio]{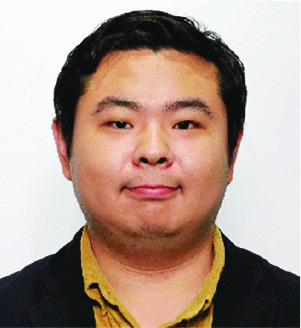}}]{Botian Shi}
received the Ph.D. degree from the School of Computer Science \& Technology, Beijing Institute of Technology, China, in 2021. He is currently a Researcher at ADLab of Shanghai Artificial Intelligence Laboratory, Shanghai, China. His research interests include Autonomous Driving Systems, Embodied Artificial Intelligence as well as Knowledge-driven Autonomous Driving.
\end{IEEEbiography}


\begin{IEEEbiography}[{\includegraphics[width=1in,height=1.25in,clip,keepaspectratio]{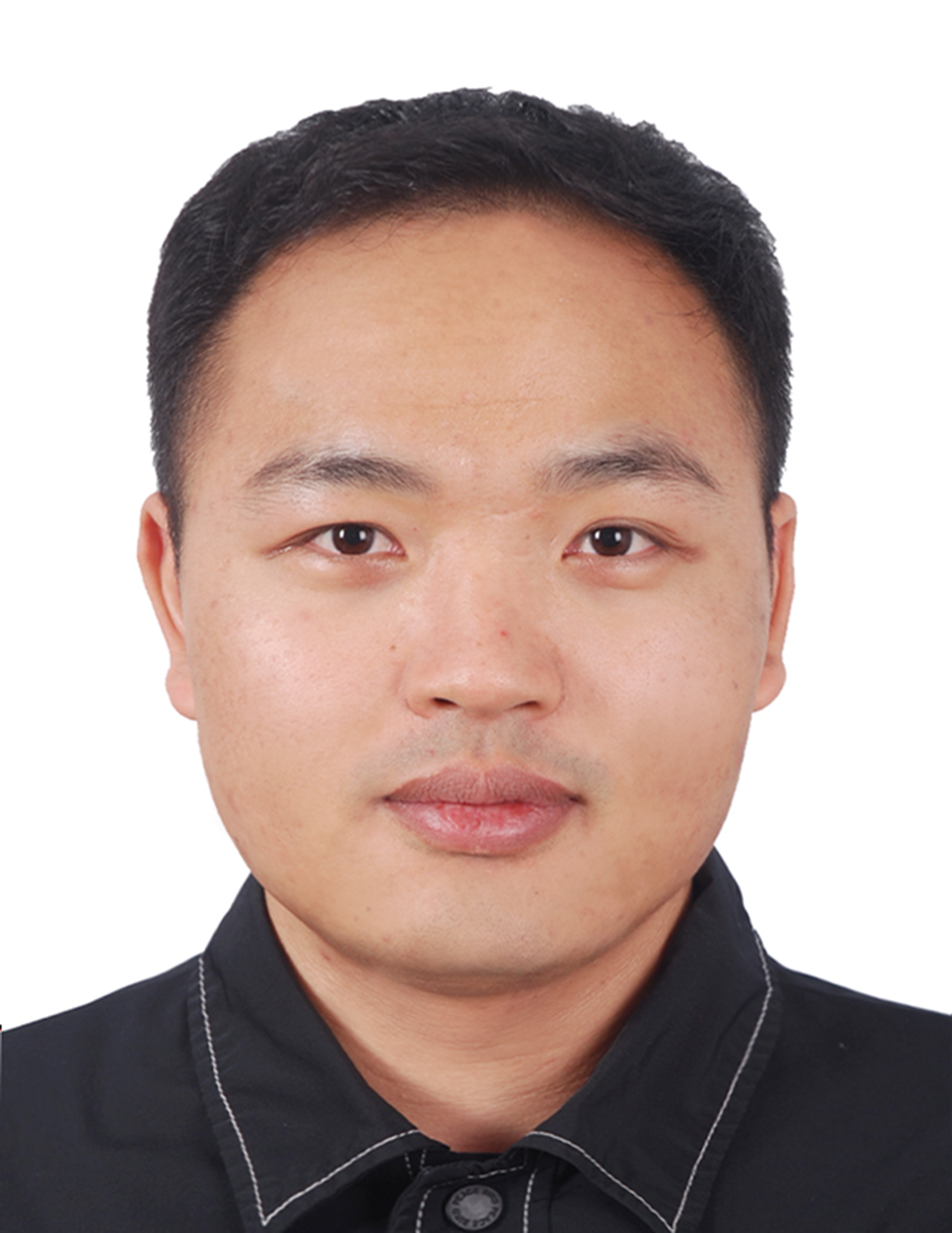}}]{Wenqi Shao}
is a Research Scientist at Shanghai Artificial Intelligence Laboratory. He completed his PhD in 2022 at the Multimedia Lab of the Chinese University of Hong Kong (CUHK). Prior to his doctoral studies, he obtained a bachelor's degree from School of Mathematics at the University of Electronic Science and Technology of China (UESTC) in 2017.
His research interests revolve around multimodal foundation models, large language model compression,  transfer learning, and applications in multimedia.
\end{IEEEbiography}


\begin{IEEEbiography}[{\includegraphics[width=1in,height=1.25in,clip,keepaspectratio]{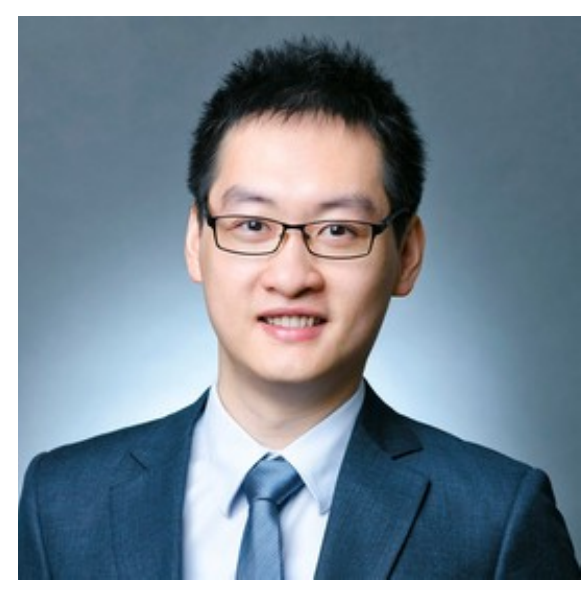}}]{Ping Luo} is an Associate Professor in the department of computer science, The University of Hong Kong (HKU). He received his PhD degree in 2014 from Information Engineering, the Chinese University of Hong Kong (CUHK), supervised by Prof. Xiaoou Tang and Prof. Xiaogang Wang. He was a Postdoctoral Fellow in CUHK from 2014 to 2016. He joined SenseTime Research as a Principal Research Scientist from 2017 to 2018. His research interests are machine learning and computer vision. He has published 100+ peer-reviewed articles in top-tier conferences and journals such as TPAMI, IJCV, ICML, ICLR, CVPR, and NIPS. His work has high impact with 78000+ citations according to Google Scholar. He has won a number of competitions and awards such as the first runner up in 2014 ImageNet ILSVRC Challenge, the first place in 2017 DAVIS Challenge on Video Object Segmentation, Gold medal in 2017 Youtube 8M Video Classification Challenge, the first place in 2018 Drivable Area Segmentation Challenge for Autonomous Driving, 2011 HK PhD Fellow Award, and 2013 Microsoft Research Fellow Award (ten PhDs in Asia).
\end{IEEEbiography}


\begin{IEEEbiography}[{\includegraphics[width=1in,height=1.25in,clip,keepaspectratio]{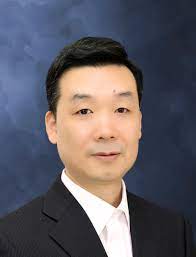}}]{Yu Qiao} (Senior Member, IEEE) received the Ph.D. degree from The University of Electro-Communications, Japan, in 2006.
He now is the lead scientist at Shanghai AI Laboratory, a researcher, and the honored director of the Multimedia Laboratory at Shenzhen Institutes of Advanced Technology at the Chinese Academy of Sciences (SIAT).
He served as an assistant professor at the Graduate School of Information Science and Technology at the University of Tokyo from 2009 to 2010. He has been working on deep learning since 2006, and he is one of the earliest people to introduce deep learning to video understanding. He and his team invented center loss and temporal segment networks. He has published more than 400 articles in top-tier conferences and journals and conferences in computer science with more than 60,000 citations. He received the CVPR 2023 Best Paper Award and AAAI 2021 Outstanding Paper Award. His research interests revolve around foundation models, computer vision, deep learning, robotics, and AI applications.
\end{IEEEbiography}


\begin{IEEEbiography}[{\includegraphics[width=1in,height=1.25in,clip,keepaspectratio]{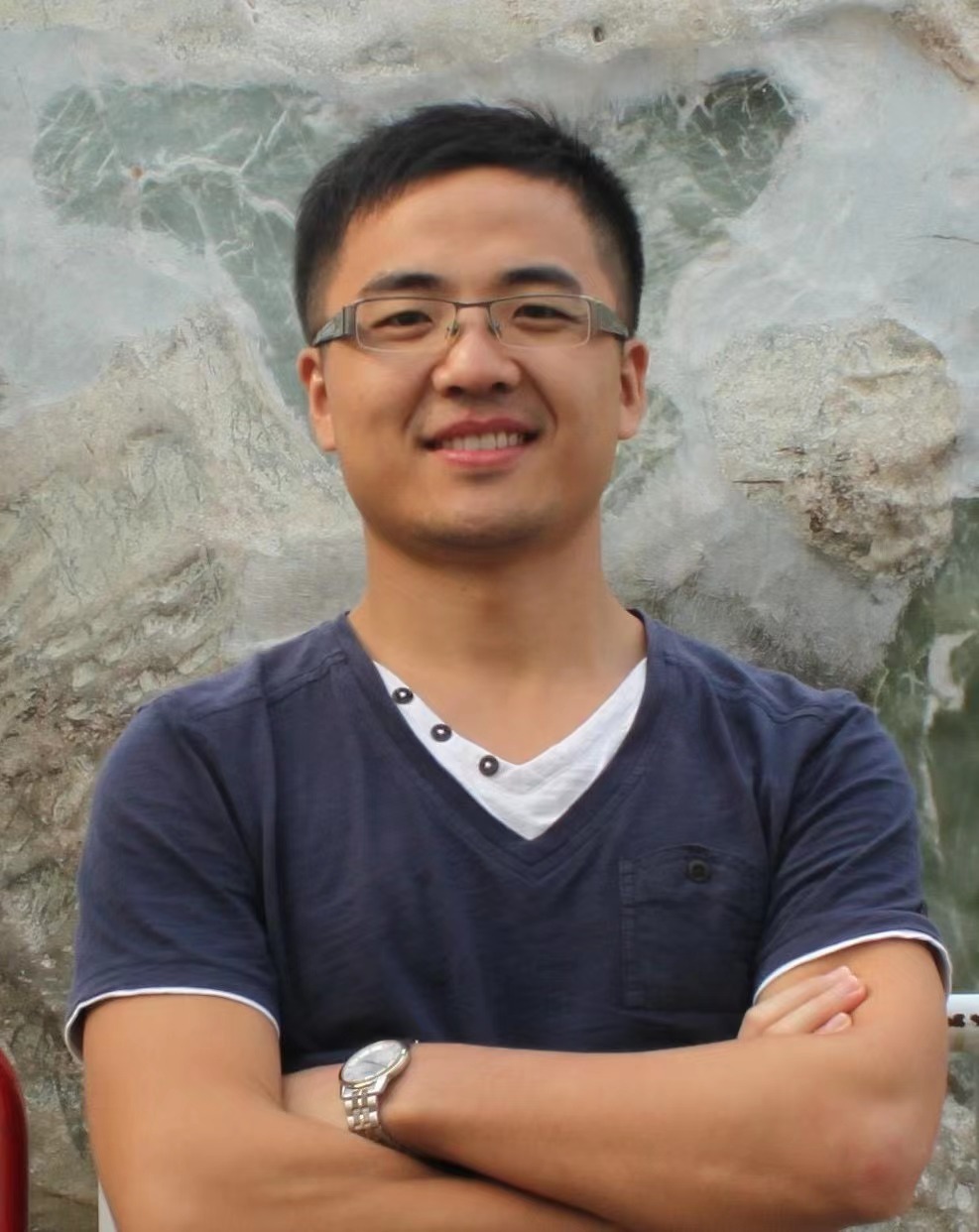}}]{Tao Chen} (Senior Member, IEEE)
received the Ph.D. degree in information engineering from Nanyang Technological University, Singapore, in 2013. He was a Research Scientist at the Institute for Infocomm Research, A*STAR, Singapore, from 2013 to 2017, and a Senior Scientist at
the Huawei Singapore Research Center from 2017 to 2018. Since 2019, he joined Fudan and led a research team focusing on light deep vision model design, multimodal vision analysis, and edge device-aware vision applications. To date, Dr. Tao Chen has undertaken multiple projects and fundings from various government agencies such as NSFC and corporations like Tencent. He has published over 110 academic papers in various reputable journals and conferences like IEEE T-PAMI/IJCV/T-IP/CVPR/NeurIPS, etc., and has granted over 10 PCT patents. 
\end{IEEEbiography}

\begin{IEEEbiography}[{\includegraphics[width=1in,height=1.25in,clip,keepaspectratio]{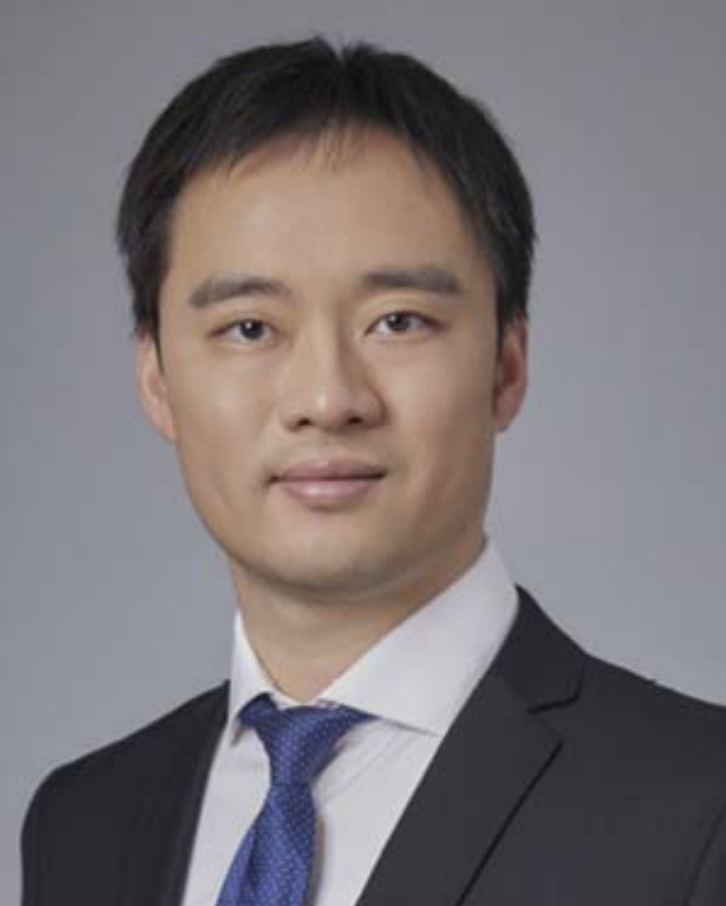}}]{Junchi Yan} (Senior Member, IEEE) is the Deputy `Director and Professor with School of Artificial Intelligence and Department of Computer Science and Engineering, Shanghai Jiao Tong University (SJTU), Shanghai, China. Before that, he was a Research Staff Member with IBM Research where he started his career since April 2011 and until 2018. He obtained his Ph.D. in Electronic Engineering with SJTU in 2015. He received the Best Paper Candidate of CVPR 2024. His research interests include machine learning and applications. He regularly serves as Senior PC/Area Chair for NeurIPS, ICML, ICLR, CVPR, AAAI, IJCAI, SIGKDD, and Associate Editor for IEEE TPAMI, Pattern Recognition. He is a Fellow of IAPR.
\end{IEEEbiography}

\end{document}